\documentclass[a4paper,fleqn]{cas-dc}
\usepackage[numbers]{natbib}

\def\tsc#1{\csdef{#1}{\textsc{\lowercase{#1}}\xspace}}
\tsc{WGM}
\tsc{QE}

\begin{document}
\let\WriteBookmarks\relax
\def\floatpagepagefraction{1}
\def\textpagefraction{.001}
\let\printorcid\relax 

\shorttitle{}    

\shortauthors{Yi~Liu et al.}

\title[mode = title]{Seamless Detection: Unifying Salient Object Detection and Camouflaged Object Detection}  

\author[1]{Yi~Liu}
\author[1]{Chengxin~Li}
\author[1]{Xiaohui~Dong}
\author[2]{Lei~Li}
\author[3]{Dingwen~Zhang}
\cormark[1]
\author[1]{Shoukun~Xu}
\author[4]{Jungong~Han}
\cormark[1]

\address[1]{School of Computer Science and Artificial Intelligence, Changzhou University, Changzhou 213159, China} 
\address[2]{Science and Technology on Complex System Control and Intelligent Agent Cooperation Laboratory, Beijing, China} 
\address[3]{School of Automation, Northwestern Polytechnical University, Xi’an 710129, China} 
\address[4]{Department of Automation,  Tsinghua University, Beijing 100084, China} 
\cortext[1]{Equally Corresponding authors}  

\begin{abstract}
Achieving joint learning of Salient Object Detection (SOD) and Camouflaged Object Detection (COD) is extremely challenging due to their distinct object characteristics, \emph{i.e.}, saliency and camouflage. The only preliminary research treats them as two contradictory tasks, training models on large-scale labeled data alternately for each task and assessing them independently. However, such task-specific mechanisms fail to meet real-world demands for addressing unknown tasks effectively. To address this issue, in this paper, we pioneer a task-agnostic framework to unify SOD and COD. To this end, inspired by the agreeable nature of binary segmentation for SOD and COD, we propose a \textbf{C}ontrastive \textbf{D}istillation \textbf{P}aradigm (\textbf{CDP}) to distil the foreground from the background, facilitating the identification of salient and camouflaged objects amidst their surroundings. To probe into the contribution of our CDP, we design a simple yet effective contextual decoder involving the interval-layer and global context, which achieves an inference speed of \textbf{67 fps}. Besides the supervised setting, our CDP can be seamlessly integrated into unsupervised settings, eliminating the reliance on extensive human annotations. Experiments on public SOD and COD datasets demonstrate the superiority of our proposed framework in both supervised and unsupervised settings, compared with existing state-of-the-art approaches. Code is available on {\color{blue}https://github.com/liuyi1989/Seamless-Detection.}
\end{abstract}




\maketitle

\section{Introduction}
Visually Salient Object Detection (SOD) and Camouflaged Object Detection (COD), which are fundamental image segmentation tasks in the computer vision community, have garnered significant research attention. SOD targets detecting attracting-attention objects that are usually stand out from their surroundings, while COD aims to discover objects that are usually concealed in their surroundings \cite{liu2021integrating}. Due to the object characteristics of SOD and COD, they have been embedded separately in large-scale applications, such as SOD for adversarial defense \cite{li2023revisiting, yu2021salience}, and COD for traffic classification \cite{zhang2024enhanced} and surveillance \cite{fang2023surveillance}. 

\begin{figure}[t]
	\centering
	\includegraphics[width=0.92\linewidth]{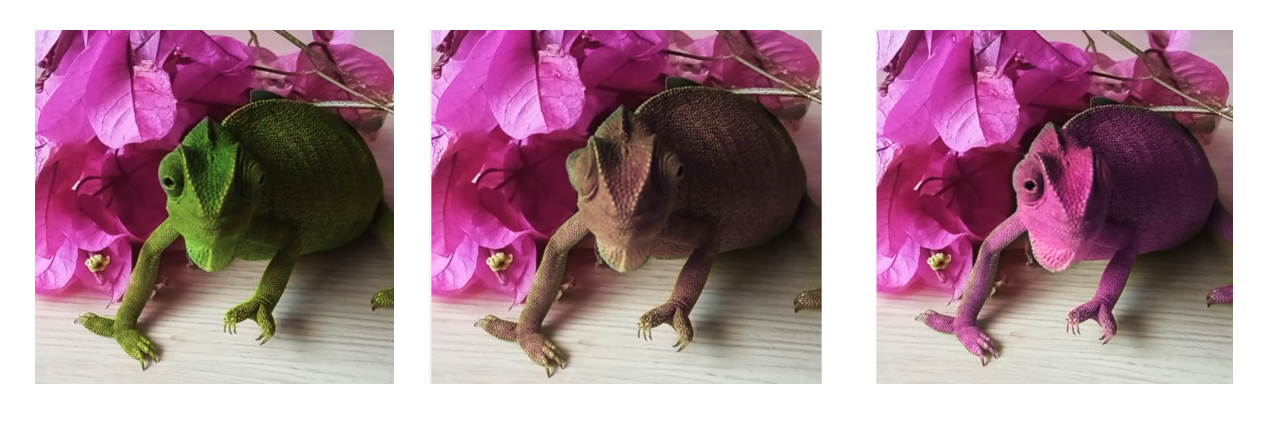}
	\caption{An easy example for the co-existing saliency and camouflage scene. A chameleon is salient when it appear in a new scene. However, it will change its body appearance to conceal itself in the surroundings, which makes it camouflaged. For the event, it is not reasonable to detect the chameleon using the individual salient or camouflaged object detection model. Inspired by this observation, it is necessary to design a task-agnostic model unifying the abilities of saliency and camouflaged detection.}
	\label{fig: s_c_scene}
\end{figure}

In real-world scenarios, objects often exhibit multiple characteristics simultaneously, such as saliency and camouflage. As shown in Fig. \ref{fig: s_c_scene}, a chameleon is salient when it appear in a new scene. However, it will change its body appearance to conceal itself in the surroundings, which makes it camouflaged. These qualities play crucial roles in various practical applications, including autonomous driving where detecting both salient features and camouflaged objects is essential for safety \cite{lateef2021saliency, song2022msfanet}, and in remote sensing where identifying saliency and camouflage aids in data analysis \cite{diao2016efficient, dehmollaian2006electromagnetic}. Despite significant advancements in separate SOD and COD techniques, they are confined to specific tasks. Since the object characteristics cannot be forewarned in real-world life, the current separate SOD and COD methods will fail in the opposite tasks. For example, it is not reasonable to detect the chameleon in Fig. \ref{fig: s_c_scene} using the individual salient or camouflaged object detection model. Although \cite{li2021uncertainty} makes the joint learning of SOD and COD, its training and testing are still task-specific, which cannot solve the un-forewarned case. To solve this urgent problem, in this paper, \textit{we focus on the task-agnostic unified framework of SOD and COD towards the real-world practicability}.
\begin{figure*}[t]
	\centering
	\includegraphics[width=0.88\linewidth]{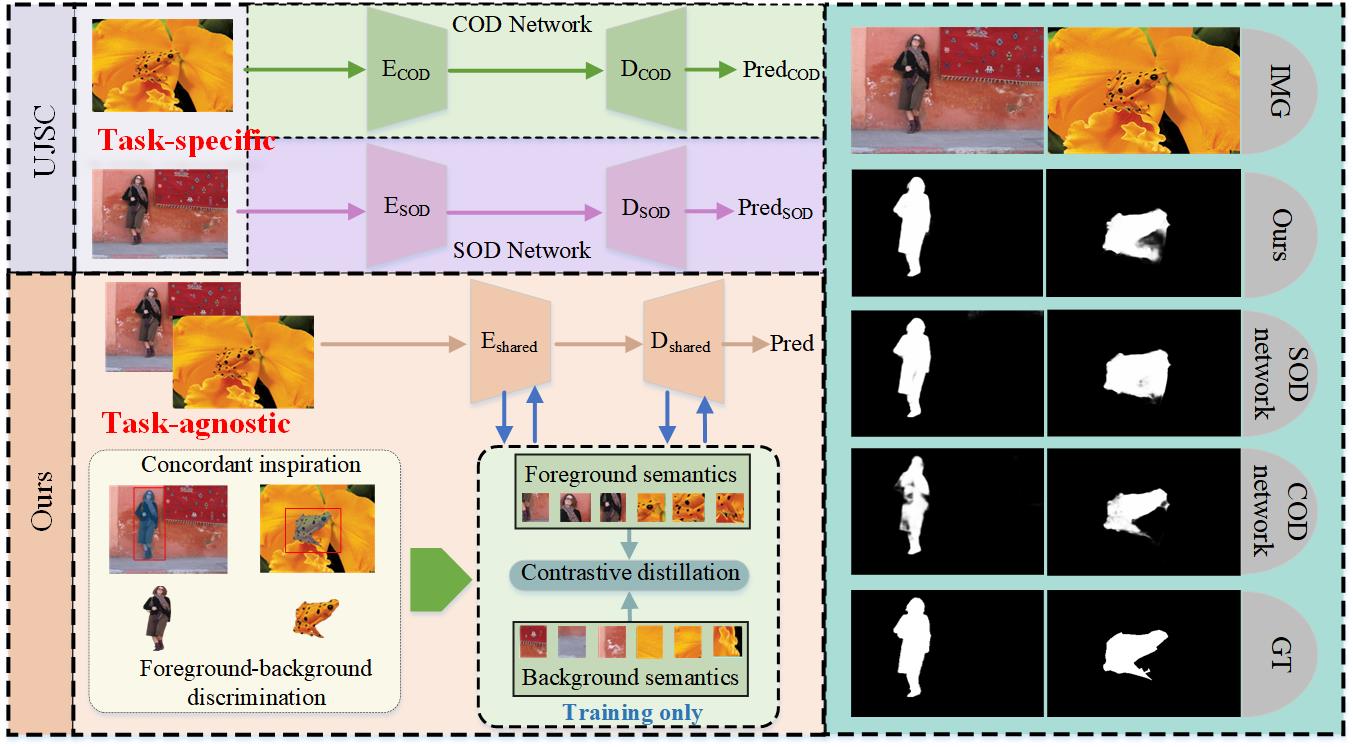}
	\caption{Motivation statement. The previous UJSC \cite{li2021uncertainty} is task-specific, which must fed salient and camouflaged image into SOD network and COD network correspondingly, otherwise generating poor results. This challenge can well be solved by our task-agnostic framework.}
	\label{fig: illustration}
\end{figure*}

Thanks to the large-scale pixel-level annotated benchmarks, \textit{e.g.}, DUTS \cite{wang2017learning} and COD10K \cite{fan2020camouflaged}, deep supervised research in both SOD and COD has advanced significantly in separate avenues. However, the joint learning of SOD and COD is still in its infancy. The work \cite{li2021uncertainty}, named UJSC, makes specialized study towards this issue via treating SOD and COD as two contradictory tasks, which is implemented by two branches of encoders and decoders that share the structures but not parameters. As shown in Fig. \ref{fig: illustration}, the procedure of UJSC \cite{li2021uncertainty} can be described as i) Besides the SOD image and COD image in the training stage, extra general object images are required for the contradiction measure to regularize SOD and COD encoders; ii) UJSC \cite{li2021uncertainty} manually labels SOD as 0 and COD as 1 to train the uncertainty learning separately; iii) UJSC \cite{li2021uncertainty} trains SOD and COD branches using SOD and COD datasets separately in an alternate manner; iv) UJSC \cite{li2021uncertainty} infers the SOD and COD task separately. While these steps represent significant progress for UJSC \cite{li2021uncertainty}, they make UJSC \cite{li2021uncertainty} a task-specific model, failing to meet the task-agnostic requirement in the real-world unforewarned case. 

In this paper, we make a preliminary exploration into the task-agnostic unified framework for SOD and COD. Traditionally, human has the ability to identify the target from its surroundings, irrespective of whether they are salient or camouflaged, highlighting an agreeable nature of SOD and COD: the identification of foreground from background. Inspired by such agreeable nature of binary segmentation, we propose a \textbf{C}ontrastive \textbf{D}istillation \textbf{P}aradigm (\textbf{CDP}) to distil the target from the background. As shown in our framework of Fig. \ref{fig: illustration}, the SOD and COD tasks share the same encoder and decoder in terms of structures and parameters, involving one input for mixed SOD \& COD images. During training, the foreground semantics and background semantics, derived from the decoder and encoder, respectively, are fed into the contrastive distillation framework, which is learned by the contrastive loss. During test, both SOD and COD tasks are inferred using the same encoder and decoder with the same parameters. Compared with UJSC \cite{li2021uncertainty}, our CDP has three advantages: i) No extra general object images avoid the contradiction measure; ii) Only one input for mixed SOD \& COD images ensures the proposed framework task-agnostic in the training and testing stages, making it adaptable to real-world applications; iii) Our CDP can be plugged in the existing image segmentation models. To discuss the effectiveness of our CDP, we design simple but effective encoder and decoder, which produces a real-time inference with 67 fps. To be concrete, we opt for ResNet-50 \cite{he2016deep} as our encoder choice. The decoder is designed using the integration of \textbf{I}nterval-layer and \textbf{G}lobal \textbf{C}ontext (\textbf{IGC}) in which the interval layers and the deepest layer of the backbone are involved to be integrated to decode the image semantics.

Furthermore, UJSC \cite{li2021uncertainty} tackles the joint learning of SOD and COD using large-scale pixel-level SOD and COD labeled benchmarks, demanding significant human effort, with each image label consuming approximately 60 minutes \cite{fan2020camouflaged}. Although there have been a lot of attempts for unsupervised SOD \cite{zhang2017supervision, wang2022multi, lin2022causal, zhou2022activation, nguyen2019deepusps, shin2022unsupervised, melas2022deep, zhou2023texture} and unsupervised COD \cite{zhang2023unsupervised} separately, the joint unsupervised learning of SOD and COD is an undeveloped field, which is a demand in real-world life. Due to its versatility, our CDP extends beyond the confines of supervised learning, enabling the joint unsupervised learning of SOD and COD. Concretely, the deep features of DINO \cite{caron2021emerging} are parsed to generate the initial pseudo mask for supervision at the first two epochs. In the following, our CDP absorbs the foreground semantics and pseudo background semantics for contrastive learning to distil the foreground objects. To train the model with high-quality labels, pseudo labels are updated for each epoch. 

Contributions of this paper are listed as follows:

i) We make the study to unify SOD and COD in a task-agnostic framework via a contrastive distillation paradigm, inspired by their agreeable nature of binary segmentation.

ii) By unifying SOD and COD on both supervised and unsupervised settings, we alleviate the need for extensive human annotations, thereby reducing the laborious task of large-scale dataset labeling.


iii) Experiments on public SOD and COD benchmarks demonstrate that our task-agnostic framework achieves the competitive performance in the supervised setting and State-Of-The-Art (SOTA) performance in the unsupervised setting, compared with the previous task-specific methods.

The paper is organized as follows. Sec. \ref{sec:Related} reviews the related work to the proposed method. Sec. \ref{sec:Proposed} introduces details of the proposed method. Sec. \ref{sec:Experiment} conducts a serious of experiments to understand the proposed method. Sec. \ref{sec:Conclusion} concludes the paper.

\section{Related work}
\label{sec:Related}

In this section, we will review the related works to our method, including supervised SOD and COD, unsupervised SOD and COD, and contrastive learning.
\subsection{Supervised SOD and COD}
Before deep learning \cite{krizhevsky2012imagenet}, SOD and COD move forward relying on the hand-crafted methods \cite{borji2015salient, liu2018salient, jiang2013salient, mondal2020camouflaged}. The emergence of deep learning has broken the performance bottleneck of these tasks, which has been a popular trend.

\textbf{SOD.} The development of deep SOD began in \cite{han2014background} that used a residual reconstruction network. Thereafter, deep learning has been largely developed for SOD \cite{wang2021salient} thanks to the large-scale pixel-level labels, using Muli-Layer Perceptron (MLP) classifiers \cite{li2015visual, zhao2015saliency}, Fully Convolutional Network (FCN) \cite{siris2021scene, pang2020multi, zhao2019egnet, chen2018reverse, wang2018salient, tian2023modeling}, Capsule Networks (CapsNets) \cite{liu2019employing, liu2021part, liu2022disentangled}, and Transformer \cite{liu2021visual, zhang2021learning}.

\textbf{COD.} Inspired by the hunting of predators, Fan \textit{et al.} \cite{fan2020camouflaged} designed a search-identification network to detect the camouflaged object, which turns on the research on deep COD. Thereafter, a lot of attempts have been devoted to the development of deep COD. For example, Zhai \textit{et al.} \cite{zhai2021mutual} used graph learning towards COD. Zhong \textit{et al.} \cite{zhong2022detecting} detected the camouflaged object using the frequency domain knowledge. Pang \textit{et al.} \cite{pang2022zoom} introduced the multi-scale detection network for COD. Huang \textit{et al.} \cite{huang2023feature} designed a feature shrinkage pyramid architecture using Transformer to detect the camouflaged object. He \textit{et al.} \cite{he2023camouflaged} introduced the learnable wavelets towards the task of COD.

There are efforts towards the joint learning of SOD and COD. Li \textit{et al.} \cite{li2021uncertainty} implemented the joint learning of SOD and COD via learning the task contradiction and uncertainty. However, their task-specific model cannot solve the real-world unforewarned case, which can be solved well by our task-agnostic framework. 
\subsection{Unsupervised SOD and COD}
Traditional deep SOD and COD usually rely on large-scale pixel-level labels, \textit{e.g.}, DUTS \cite{wang2017learning} and COD10K \cite{fan2020camouflaged}, which consume huge labour. To solve this problem, unsupervised learning without human annotations is employed to implement SOD and COD.

\textbf{Unsupervised SOD.} Zhang \textit{et al.} \cite{zhang2017supervision} opened the research of unsupervised SOD, which generated high-quality pseudo labels via discovering consistency from noisy traditional detectors \cite{shi2015hierarchical, zhang2015minimum, zhang2015exploiting}. Following this route, there are a few works towards unsupervised SOD \cite{nguyen2019deepusps, zhang2018deep}. More recently, unsupervised SOD generates high-quality pseudo labels from high-level deep semantics. For example, Zhou \textit{et al.} \cite{zhou2022activation} activated the multi-level semantics for high-quality labels generation to train the detector. Later on, they developed unsupervised SOD via mining saliency knowledge from easy and hard samples \cite{zhou2023texture}. Shin \textit{et al.} \cite{shin2022unsupervised} introduced spectral clustering to generate pseudo labels for unsupervised SOD.

\textbf{Unsupervised COD.} In \cite{zhang2023unsupervised}, a source-free unsupervised domain adaptation was introduced to solve the task of unsupervised COD.

There have few efforts to the joint unsupervised learning of SOD and COD. In this paper, we make the study for unified SOD and COD without human annotations.
\subsection{Contrastive learning}
Recently, contrastive learning, which learns general and robust feature representations by comparing similar and dissimilar pairs, has shown power in a lot of computer vision tasks. For example, Lo \textit{et al.} \cite{lo2021clcc} learned better illuminant-dependent features for color constancy via constructing contrastive pairs. Li \textit{et al.} \cite{li2022targeted} devised a targeted supervised contrastive learning framework to enhance the feature distribution uniformity for further image recognition in the case of long-tail data. Tang \textit{et al.} \cite{tang2022contrastive} segmented point cloud objects using a contrastive boundary learning framework. Zhao \textit{et al.} \cite{zhao2021contrastive} achieved semantic segmentation with limited labels using a contrastive training strategy. Zhang \textit{et al.} \cite{zhang2023clamp} solved the cross-modal animal pose estimation problem using a contrastive learning paradigm for the language knowledge and animal pose images. Meng \textit{et al.} \cite{meng2023tracking} combined the learning of self-contrast, cross-contrast, and ambiguity contrast for multi-object tracking.

In this paper, we used contrastive learning to make SOD and COD tasks in a unified paradigm via distilling the target from the background.
\begin{figure*}[t]
	\centering
	\includegraphics[width=0.96\linewidth]{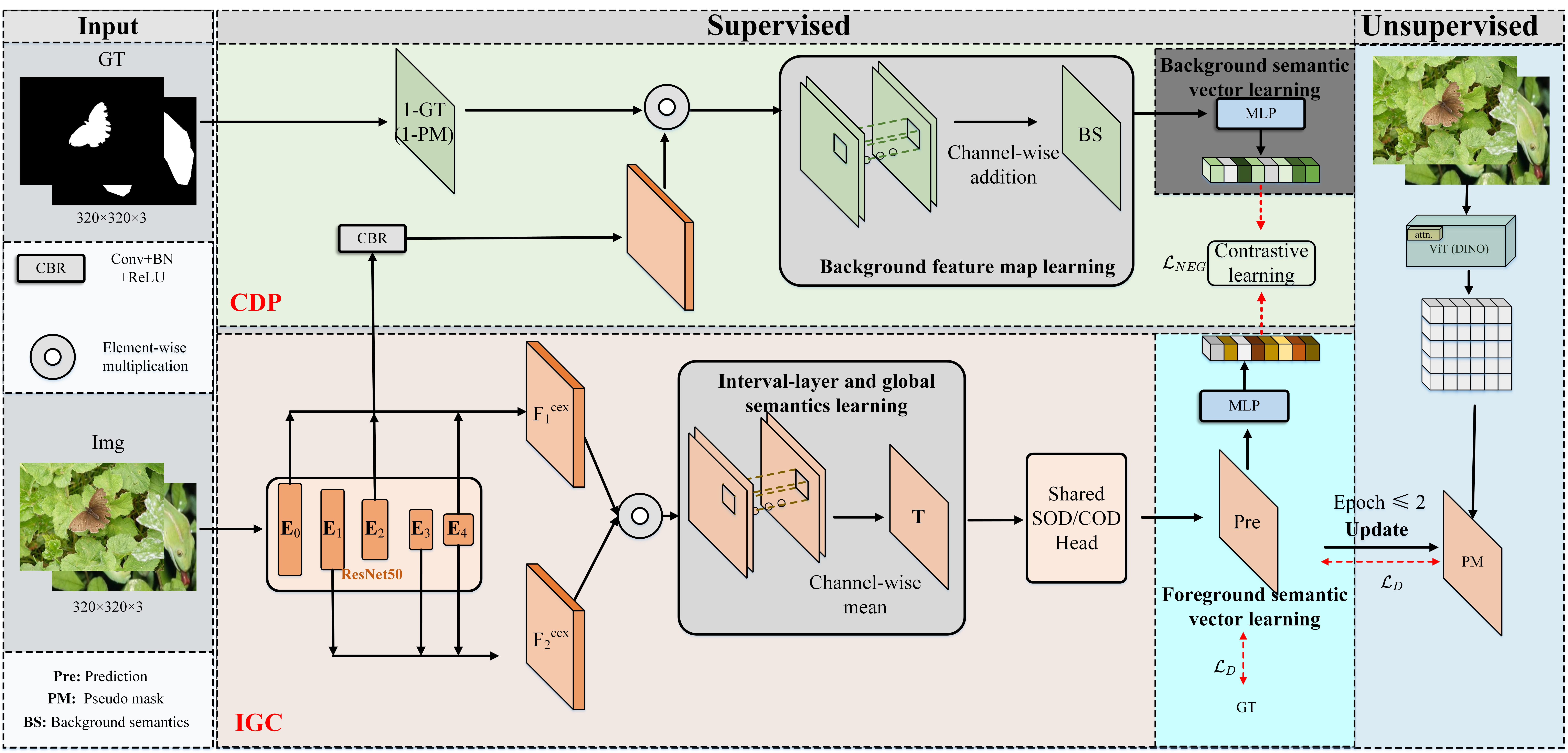}
	\caption{Overview of the framework. $\textbf{E}_\ast$ is the last layer of different blocks in ResNet-50 \cite{he2016deep}. ${\cal L}_D$ and ${\cal L}_{NEG}$ denote the training losses of Eq. (\ref{Dloss}) and Eq. (\ref{loss: contr}), respectively. Under the supervised setting, the foreground map inferred by IGC is supervised by ground truth. Besides, the foreground semantics and background semantics, generated by the  decoder and encoder of IGC, respectively, are supervised using the contrastive loss within CDP. Under the unsupervised setting, the deep features of DINO \cite{caron2021emerging} are parsed to generate the pseudo masks at the initial two epochs, which will be updated at each epoch. Note that, only IGC is run for inference at the test stage for both supervised and unsupervised settings.}
	\label{framework}
\end{figure*}

\section{Proposed method}
\label{sec:Proposed}

In this section, we will illustrate the proposed method with clear details for techniquely understanding. 
\subsection{Overview}
Fig. \ref{framework} overviews the proposed framework. It includes supervised and unsupervised settings. In the supervised setting, IGC infers the foreground map, which is supervised by ground truth. Besides, the foreground semantics and background semantics, generated by the decoder and encoder of IGC, respectively, are supervised using the contrastive loss within CDP. Under the unsupervised setting, The deep features of DINO \cite{caron2021emerging} are parsed to generate the initial pseudo mask for supervision at the first two epochs, which will be updated at each epoch. IGC is run for inference at the test stage for both supervised and unsupervised settings.
\subsection{Interval-layer and global contextual network}

Context is an important semantic in deep neural networks. To design a simple but superior network to decode the image semantics for further inference, we dig into the context implicitly contained in the deep backbone, including interval-layer context and global context.

Suppose the image denoted as ${\bf{I}} \in {\Re ^{H \times W \times 3}}$, which is fed into ResNet-50 \cite{he2016deep} to extract multi-level deep features, denoted as ${{\bf{E}}_i} \in {\Re ^{\frac{H}{{{2^{i + 1}}}} \times \frac{W}{{{2^{i + 1}}}} \times C_i}},i \in [0,1,2,3,4], C_i \in [64, 256, 512, 1024, 2048]$. First, the interval-layer features, \textit{i.e.}, ${\textbf{E}}_0$ \& ${\textbf{E}}_2$ and ${\textbf{E}}_1$ \& ${\textbf{E}}_3$, have different receptive fields with respect to the input image, concretely $2^2 \times$ spatial difference. These interval-layer contexts will help to capture different-scale objects. Besides, the deepest-layer features, \textit{i.e.}, ${\textbf{E}}_4$ learns the high-level image semantics from the global perspective thanks to the $2^5 \times$ receptive field. Based on these two findings, we expect to integrate the interval-layer contexts and the global context for the semantic exploration of the input image. Specifically, All ${\textbf{E}}_i$s are transformed into the same-channel versions, \textit{i.e.},
\begin{equation}
	{{\textbf{F}}_i} \in {\Re ^{\frac{H}{{{2^{i + 1}}}} \times \frac{W}{{{2^{i + 1}}}} \times 64}} = f_{conv}({{\textbf{E}}_i}),i = 0,1,2,3,
	\label{equ: Channel}
\end{equation}
where $f_{conv}\left(  \cdot \right)$ means the operation of convolution.

On top of that, the interval-layer features, \textit{i.e.}, ${\textbf{F}}_0$ \& ${\textbf{F}}_2$ and ${\textbf{F}}_1$ \& ${\textbf{F}}_3$, and the global features, \textit{i.e.}, ${\textbf{F}}_4$ are mixed together to combine the interval-layer context and the global context, \textit{i.e.},
\begin{equation}
	\begin{array}{l}
		{{\textbf{F}}^{cex}_1}  \in {\Re ^{\frac{H}{{{2^{2}}}} \times \frac{W}{{{2^{2}}}} \times 64}} = f_{conv}({\left[\kern-0.15em\left[ {{\textbf{F}}_1}, bi({{\textbf{F}}_3}), bi({{\textbf{F}}_4}))\right]\kern-0.15em\right]}),\\
		{{\textbf{F}}^{cex}_2} \in {\Re ^{\frac{H}{{{2^{1}}}} \times \frac{W}{{{2^{1}}}} \times 64}} = f_{conv}({\left[\kern-0.15em\left[ {{\textbf{F}}_0}, bi({{\textbf{F}}_2}), bi({{\textbf{F}}_4}))\right]\kern-0.15em\right]}),\\
	\end{array}
	\label{equ: integration}
\end{equation}
where ${\left[\kern-0.15em\left[ *\right]\kern-0.15em\right]}$ means the concatenation operation along the channel dimension. $bi(\cdot)$  denote the upsampling operations using the bilinear interpolation.

In the following, the context-aware features ${{\textbf{F}}^{cex}_1}$ and ${{\textbf{F}}^{cex}_2}$ are integrated further to compute the final context semantics, \textit{i.e.},
\begin{equation}
	\begin{array}{l}
		{{\textbf{F}}^{cex}} \in {\Re ^{\frac{H}{{{2^{1}}}} \times \frac{W}{{{2^{1}}}} \times 64}} = bi({{\textbf{F}}^{cex}_1})  \odot  {{\textbf{F}}^{cex}_2},
	\end{array}
	\label{equ: ele-mul}
\end{equation} 
where $\odot$ means the element-wise multiplication.

The target inference is obtained by an average along the channel dimension, \textit{i.e.},
\begin{equation}
	\begin{array}{l}
		{{\textbf{T}}} \in {\Re ^{\frac{H}{{{2^{1}}}} \times \frac{W}{{{2^{1}}}} \times 1}} =  \frac{{\sum\nolimits_{i = 1}^{64} {{{\bf{F}}^{cex}}\left( {:,:,i} \right)} }}{{64}}.
	\end{array}
	\label{equ: infer}
\end{equation} 

The final inference map can be achieved by activating ${{\textbf{T}}}$ using the Sigmoid function, \textit{i.e.},
\begin{equation}
	\begin{array}{l}
		{{\textbf{T}}^{act}} \in {\Re ^ {{H} \times {W} \times 1}} = \frac{1}{{1 + {e^{ - \left(  bi({\bf{T}})\right)}}}}
	\end{array}.
	\label{equ: finalinfer}
\end{equation}

\subsection{Contrastive distillation paradigm}
To identify the targets from the surroundings, we propose a CDP model to distil the foreground from the background, which suits the saliency and camouflage scenes. The idea is to compute the contrast between the target inference and the background with the purpose of training the model for better inference. To this end, as shown in Fig. \ref{framework}, CDP is implemented by three components, including foreground semantic vector, background semantic vector, and foreground-background contrastive learning, which will be described in detail in the following.

Given a batch of $n$ samples, denoted as ${{\bf{I}}_{\{ 1:n\} }}$, the foreground semantic vector ${\bf{v}}_{\{ 1:n\} }^f$ is learned via a fully-connected layer on the inference ${\textbf{T}}$, \textit{i.e.}, 
\begin{equation}
	{\bf{v}}_{\{ 1:n\} }^f \in {\Re ^{64}} = {\bf{W}}_b{\bf{T}}+{\bf{b}}_b
	\label{equ: vf}
\end{equation}
where ${\bf{W}}_b$ and ${\bf{b}_b}$ are the learned weights and bias in the fully-connected layer.

The background semantic vector ${\bf{v}}_{\{ 1:n\} }^b$ is learned using the background mask (${1- \bf{GT}}$) and the intermediate-layer backbone features, which is determined as the third-layer features ${\bf{E}}_2$ of the input image due to their suitable receptive fields. Concretely, 
\begin{equation}
	{\bf{v}}_{\{ 1:n\} }^b \in {\Re ^{64}} = {\bf{W}}_f \left({{{\bf{E}}_{2\{ 1:n\} }} \odot \left( {1 - dw\left( {{\bf{G}}{{\bf{T}}_{\{ 1:n\} }}} \right)} \right)}\right)+{\bf{b}}_f,
	\label{equ: vb}
\end{equation}
where $dw(\cdot)$ and \textbf{GT} mean the downsampling operation and ground truth, respectively. ${\bf{W}}_f$ and ${\bf{b}}_f$ are the learned weights and bias in the fully-connected layer.

On top of the foreground semantic vector ${\bf{v}}_{\{ 1:n\} }^f$ and the background semantic vector ${\bf{v}}_{\{ 1:n\} }^b$ within the batch, the model can be trained using the contrastive loss as
\begin{equation}
	{{\cal L}_{NEG}} =  - \frac{1}{{{n^2}}}\sum\limits_{i = 1}^n {\sum\limits_{j = 1}^n {\log (1 - \frac{{\left\langle {{\bf{v}}_i^b,{\bf{v}}_i^f} \right\rangle }}{{\left\| {{\bf{v}}_i^b} \right\|\left\| {{\bf{v}}_i^f} \right\|}})} }, 
	\label{loss: contr}
\end{equation}
where $\left\langle * \right\rangle$ and $\left\| * \right\|$ indicate the inner product and matrix modulus, respectively.

\textbf{Difference to ContrastMask \cite{wang2022contrastmask}.} ContrastMask \cite{wang2022contrastmask} migrates from base to novel by sharing query vectors, in which the foreground query and background query are obtained by averaging features from partitions within a batch of object proposals accordingly. Differently, we generate the foreground vector and background vector using pseudo-labels within an image, which tend to be more semantically specific.
%
%
\begin{figure}[t]
	\centering
	\includegraphics[width=0.94\linewidth]{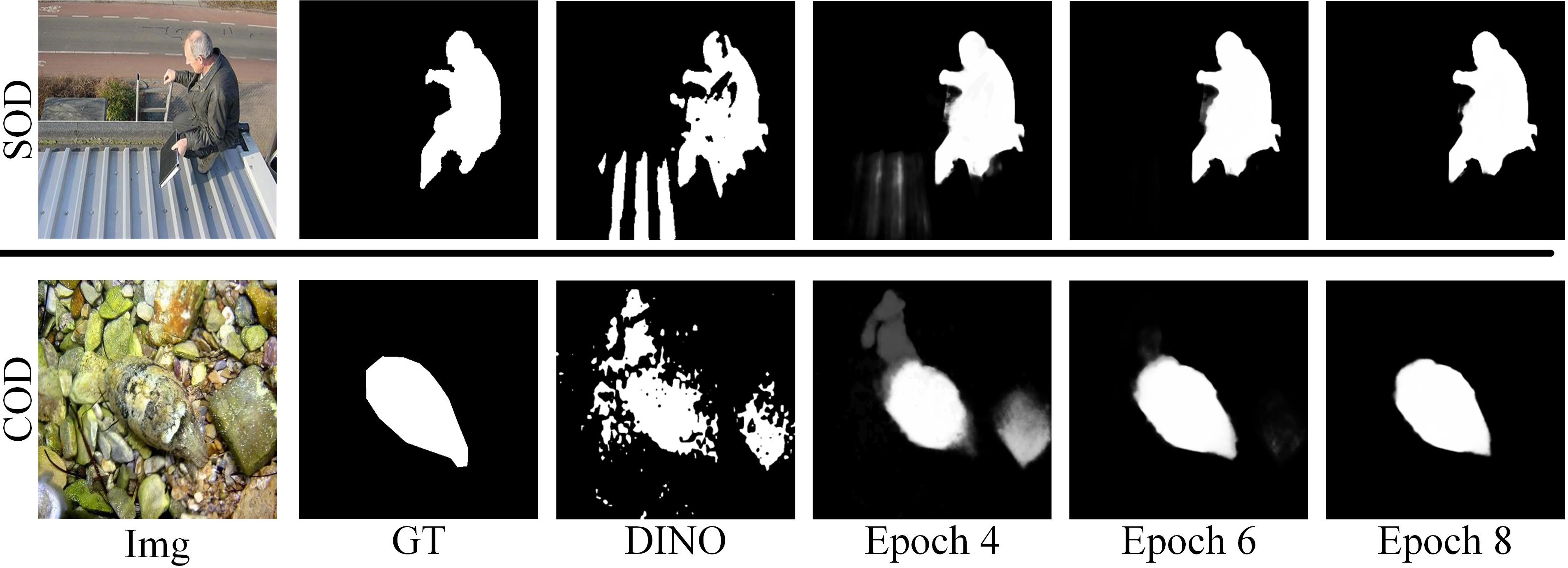}
	\caption{Visualizations for pseudo masks update.}
	\label{fig:pseudo-update}
\end{figure}
\subsection{Model training}

\subsubsection{Supervised training}
To train the model, besides the contrastive learning loss ${{\cal L}_{NEG}}$ between foreground and background, extra loss between the inference and ground truth is involved. To generate inferences closer to the ground truth in terms of three aspects, including minimum loss, structure similarity, and intersection, we take three loss functions, \textit{i.e.}, binary cross entropy loss (${{\cal L}_B}$), structural similarity loss (${{\cal L}_S}$) \cite{wang2004image}, and intersection-over-union loss (${{\cal L}_I}$), to train the model as
\begin{equation}
	{{\cal L}_D} = {{\cal L}_B}({{\textbf{T}}^{act}}, {\bf{GT}}) + {{\cal L}_S}({{\textbf{T}}^{act}}, {\bf{GT}}) + {{\cal L}_I}({{\textbf{T}}^{act}}, {\bf{GT}}).
	\label{Dloss}
\end{equation}

The overall loss ${\cal L}$ for the model learning is formulated as the combination of ${{\cal L}_{NEG}}$ and ${{\cal L}_D}$, \textit{i.e.},
\begin{equation}
	{\cal L} = {{\cal L}_{NEG}} + {{\cal L}_D}.
	\label{loss}
\end{equation}
\subsubsection{Unsupervised training}
\

\textbf{Initial pseudo mask.} Recently, DINO \cite{caron2021emerging}, a new baseline for self-supervised semantic segmentation, has been proved successful for many segmentation tasks. Inspired by its power, it is adopted to produce the initial coarse mask in our framework. Specifically, as shown in Fig. \ref{framework}, the input image is fed into the pretrained DINO \cite{caron2021emerging} to generate one class token and patch tokens. The latter are shapely transformed into dense features. On top of that, one convolution with a kernel of $1 \times 1$ is carried out to achieve the desired pseudo masks ${\bf{PM}} \in {\Re ^{^{H \times W \times 1}}}$. 

\textbf{Model training.} The loss functions of ${{\cal L}_{NEG}}$ and ${{\cal L}_D}$ are both employed to train the model. However, the unsupervised setting is learned by the pseudo mask instead of ground truth, there is some minor modifications. Firstly, ${{\cal L}_D}$ computes the contrast between the current-epoch target inference and the previous-epoch background. To this end, the background semantic vector is computed by using the previous-epoch pseudo mask ${\bf{PM}}$, \textit{i.e.},
\begin{equation}
	{\bf{v}}_{\{ 1:n\} }^b \in {\Re ^{64}} = fc\left( {{{\bf{E}}_{2\{ 1:n\} }} \odot \left( {1 - down\left( {{\bf{P}}{{\bf{M}}_{\{ 1:n\} }}} \right)} \right)} \right).
	\label{equ: vb-unsupervised}
\end{equation}

Secondly, ${{\cal L}_D}$ computes the error between the inference and the pseudo mask as
\begin{equation}
	{{\cal L}_D} = {{\cal L}_B}({{\textbf{T}}^{act}}, {\bf{PM}}) + {{\cal L}_S}({{\textbf{T}}^{act}}, {\bf{PM}}) + {{\cal L}_I}({{\textbf{T}}^{act}}, {\bf{PM}}).
	\label{Dloss-unsupervised}
\end{equation}

\textbf{Pseuod mask update.} To train the model well, pseudo labels are updated during the training epochs using the following moving average strategy
\begin{equation}
	{\bf{P}}{{\bf{M}}^i} = \left\{ \begin{array}{l}
		{\bf{P}}{{\bf{M}}^0},{\kern 1pt} {\kern 1pt} {\kern 1pt} {\kern 1pt} {\kern 1pt} {\kern 1pt} {\kern 1pt} {\kern 1pt} {\kern 1pt} {\kern 1pt} {\kern 1pt} {\kern 1pt} {\kern 1pt} {\kern 1pt} {\kern 1pt} {\kern 1pt} {\kern 1pt} {\kern 1pt} {\kern 1pt} {\kern 1pt} {\kern 1pt} {\kern 1pt} {\kern 1pt} {\kern 1pt} {\kern 1pt} {\kern 1pt} {\kern 1pt} {\kern 1pt} {\kern 1pt} {\kern 1pt} {\kern 1pt} {\kern 1pt} {\kern 1pt} {\kern 1pt} {\kern 1pt} {\kern 1pt} {\kern 1pt} {\kern 1pt} {\kern 1pt} {\kern 1pt} {\kern 1pt} {\kern 1pt} {\kern 1pt} {\kern 1pt} {\kern 1pt} {\kern 1pt} {\kern 1pt} {\kern 1pt} {\kern 1pt} {\kern 1pt} {\kern 1pt} {\kern 1pt} {\kern 1pt} {\kern 1pt} {\kern 1pt} {\kern 1pt} {\kern 1pt} {\kern 1pt} {\kern 1pt} {\kern 1pt} {\kern 1pt} {\kern 1pt} {\kern 1pt} {\kern 1pt} {\kern 1pt} {\kern 1pt} {\kern 1pt} {\kern 1pt} {\kern 1pt} {\kern 1pt} {\kern 1pt} {\kern 1pt} {\kern 1pt} {\kern 1pt} {\kern 1pt} {\kern 1pt} {\kern 1pt} i \le 2,\\
		\lambda {\bf{P}}{{\bf{M}}^i} + \left( {1 - \lambda } \right){{\bf{T}}^{act}}^{^i},{\kern 1pt} {\kern 1pt} {\kern 1pt} {\kern 1pt}   i > 2,
	\end{array} \right.
	\label{update}
\end{equation}
where the superscript means the training epoch. ${\bf{PM}}^0$ is the pseudo mask ${\bf{PM}}$ generated by parsing deep features of DINO \cite{caron2021emerging}. $\lambda$ is experimentally set to 0.4 in this paper. As shown in Fig. \ref{fig:pseudo-update}, the pseudo-mask quality improves along the training epochs.

\section{Experiment and Analysis}
\label{sec:Experiment}

In this section, we will conduct a serious of ablations and experiments to understand the contributions and performance of the proposed method.
\subsection{Implementation details}
\subsubsection{Training details} 
We use the SGD optimizer \cite{bottou2012stochastic} with an initial learning rate of 0.005 to train our model with the the batchsize of 20 using one RTX 3090 Ti GPU. The input is resized to $320 \times 320$. The training convergence occurs for 50 epochs. The training datasets include SOD training datasets (DUTS-TR training subset \cite{wang2017learning} and MSRA-B training subset \cite{liu2010learning}) and COD training datasets (CAMO \cite{le2019anabranch} and COD10K \cite{fan2020camouflaged}). Concretely, Table \ref{tab:training} lists the training details of ``Ours$_{M-C}$'' and ``Ours$_{D-C}$'' in Table \ref{tab:supervised} and \ref{tab:unsupervised}.
\begin{table}[htbp]
	\centering
	\caption{Training datasets details in Table \ref{tab:supervised} \& \ref{tab:unsupervised}. According with the SOTA unsupervised SOD methods, MSRA-B \cite{wang2017learning} is used for training our model in the unsupervised setting.}
	\scalebox{0.76}{
		\begin{tabular}{c|c|c}
			\toprule
			Model&Training datasets&Setting\\
			\midrule
			\midrule
			Ours$_{D-C}$&\makecell{DUTS-TR \cite{wang2017learning} \\10553 images} \& \makecell{CAMO \cite{le2019anabranch} \\1000 images} \& \makecell{COD10K \cite{fan2020camouflaged} \\3040 images}&\makecell{Supervised\& \\Unsupervised} \\
			\midrule
			Ours$_{D}$&\makecell{DUTS-TR \cite{wang2017learning} 10553 images}&Supervised\\
			\midrule
			Ours$_{M-C}$&\makecell{MSRA-B \cite{wang2017learning} \\2500 images} \& \makecell{CAMO \cite{le2019anabranch} \\1000 images} \& \makecell{COD10K \cite{fan2020camouflaged} \\3040 images}&Unsupervised\\
			\midrule
			Ours$_{M}$&\makecell{MSRA-B \cite{wang2017learning} 2500 images} &Unsupervised\\
			\bottomrule
	\end{tabular}}%
	\label{tab:training}%
\end{table}%
\begin{table*}[htbp]
	\centering
	\large
    \setlength{\tabcolsep}{3pt} 
	\caption{Performance ($\%$) of different methods on SOD and COD benchmarks under the supervised setting. ``--'' represents the authors have not released the results.}

	\scalebox{0.50}{\begin{tabular}{c|c|c|c|c|c|c|c|c|c|c|c|c|c}
            
			\toprule
			\multicolumn{3}{c|}{\multirow{2}[4]{*}{}} & \multicolumn{5}{c|}{SOD methods}      & \multicolumn{4}{c|}{COD methods} & \multicolumn{2}{c}{Joint learning methods} \\
			\cmidrule{4-14}    \multicolumn{3}{c|}{} & PiCANet \cite{liu2018picanet} & CPD \cite{wu2019cascaded}   & BASNet \cite{qin2019basnet} & ITSD \cite{zhou2020interactive}  & MINet \cite{pang2020multi} & SINet \cite{fan2020camouflaged} & PFNet \cite{mei2021camouflaged} & FEDER \cite{he2023camouflaged} & ZoomNet \cite{pang2022zoom} & UJSC \cite{li2021uncertainty}  & Ours$_{D-C}$  \\
			\midrule
			\multicolumn{3}{c|}{} & Task-specific&Task-specific&Task-specific&Task-specific&Task-specific&Task-specific&Task-specific&Task-specific&Task-specific& Task-specific & Task-agnostic \\
			\midrule
			\multirow{20}[10]{*}{SOD}& \multirow{4}[2]{*}{ECSSD \cite{shi2015hierarchical}} & $MAE\downarrow$   & 4.64 & 4.02 & 3.70 & 4.01 & 3.62 & 3.58 & 3.26 & 3.22 & \textbf{2.73} & 3.00 & 3.39 \\
			&   & $F_\beta \uparrow$     & 88.67 & 91.15 & 91.68 & 91.01 & 91.87 & 91.75 & 92.52 & 92.48 & 93.33          & \textbf{93.50} & 92.59 \\
			&   & $S_m \uparrow$         & 91.38 & 91.02 & 91.62 & 91.42 & 91.91 & 92.39 & 92.60 & 92.27 & \textbf{93.47} & 93.30          & 92.24 \\
			&   & $E_m \uparrow$         & 92.33 & 93.77 & 94.32 & 93.75 & 94.32 & 94.61 & 95.08 & 95.18 & 95.79          & \textbf{96.00} & 94.81 \\
			\cmidrule{2-14}
			&  \multirow{4}[2]{*}{DUTS \cite{wang2017learning}} & $MAE\downarrow$   & 5.41 & 4.29 & 4.76 & 4.23 & 3.94 & 4.10 & 3.82 & 3.92 & 3.27 & 3.20 & \textbf{3.16} \\
			&    & $F_\beta \uparrow$    & 78.20 & 82.44 & 82.24 & 83.23 & 83.49 & 82.83 & 84.31 & 84.55 &  86.63          & 86.60          & \textbf{86.65} \\
			&    & $S_m \uparrow$        & 86.70 & 86.66 & 86.56 & 87.71 & 87.49 & 87.87 & 88.17 & 87.89 &  \textbf{90.00} & 89.90          & 89.40          \\
			&    & $E_m \uparrow$        & 87.24 & 90.20 & 89.54 & 90.56 & 90.67 & 91.78 & 91.74 & 91.72 &  92.96          & \textbf{93.70} & 93.07          \\
			\cmidrule{2-14}
			&  \multirow{4}[2]{*}{DUT-O \cite{yang2013saliency}} & $MAE\downarrow$  & 6.79 & 5.67 & 5.65 & 6.32 & 5.69 & 5.63 & 5.54 & 5.63 & 5.27 & 5.10 & \textbf{4.92} \\
			&     & $F_\beta \uparrow$   & 72.24 & 73.85 & 76.68 & 75.24 & 74.04 & 75.16 & 75.99 & 77.11 & 77.12 & \textbf{78.20} & 77.43 \\
			&     & $S_m \uparrow$       & 82.64 & 81.77 & 83.62 & 82.88 & 82.18 & 83.24 & 83.25 & 83.68 & 84.09 & \textbf{85.00} & 83.86 \\
			&     & $E_m \uparrow$       & 83.28 & 84.50 & 86.50 & 85.28 & 84.58 & 85.81 & 86.24 & 87.18 & 86.61 & \textbf{88.40} & 86.85 \\
			\cmidrule{2-14}
			& \multirow{4}[2]{*}{PASCAL-S \cite{li2014secrets}} & $MAE\downarrow$   & 7.83 & 7.21 & 7.58 & 6.81 & 6.39 & 6.65 & 6.48 & 6.68 & \textbf{5.46} & -- & 5.76 \\
			&     & $F_\beta \uparrow$   & 80.02 & 82.30 & 81.77 & 83.05 & 83.03 & 82.78 & 83.33 & 82.75 & \textbf{85.05} & -- & 84.57 \\
			&     & $S_m \uparrow$       & 84.77 & 84.46 & 83.80 & 85.63 & 85.54 & 85.89 & 85.62 & 85.10 & \textbf{87.15} & -- & 86.40 \\
			&     & $E_m \uparrow$       & 86.86 & 88.25 & 87.86 & 89.15 & 89.36 & 89.22 & 89.92 & 89.26 & \textbf{91.25} & -- & 90.63 \\
			\cmidrule{2-14}
			& \multirow{4}[2]{*}{HKU-IS \cite{li2015visual}} & $MAE\downarrow$      & 4.15 & 3.32 & 3.29 & 3.46 & 3.03 & 3.21 & 2.92 & 2.92 & \textbf{2.34} & 2.60 & 2.59 \\
			&    & $F_\beta \uparrow$    & 87.08 & 89.58 & 90.36 & 89.40 & 90.55 & 89.82 & 90.47 & 90.50 & 92.33          & \textbf{92.40} & 91.87 \\
			&    & $S_m \uparrow$        & 90.54 & 90.45 & 90.77 & 90.68 & 91.39 & 91.44 & 91.44 & 91.12 & 93.08          & \textbf{93.10} & 92.16 \\
			&    & $E_m \uparrow$        & 92.26 & 94.24 & 94.30 & 93.95 & 94.65 & 94.53 & 94.94 & 94.96 & \textbf{96.14} & 86.70          & 95.58 \\
			\midrule
			\midrule
			\multirow{16}[8]{*}{COD} &  \multirow{4}[2]{*}{CAMO \cite{le2019anabranch}} & $MAE\downarrow$   & 12.50 & 11.29 & 15.90 & 10.16 & 9.03 & 9.15 & 8.49 & 7.12 & \textbf{6.59} & 7.30 & 7.17 \\
			&   & $F_\beta \uparrow$     & 57.26 & 61.77 & 47.53 & 66.29 & 69.12 & 70.20 & 74.61 & 78.09 & \textbf{79.38} & 77.17 & 76.70 \\
			&   & $S_m \uparrow$         & 70.13 & 71.61 & 61.82 & 74.99 & 74.80 & 74.54 & 78.23 & 80.21 & \textbf{81.97} & 80.01 & 79.78 \\
			&   & $E_m \uparrow$         & 71.57 & 72.27 & 66.12 & 77.99 & 79.18 & 80.35 & 84.15 & 86.66 & \textbf{87.75} & 85.87 & 84.44 \\
			\cmidrule{2-14}
			&  \multirow{4}[2]{*}{CHAMELEON \cite{skurowski2018animal}} & $MAE\downarrow$   & 8.52 & 4.80 & 11.79 & 5.73 & 3.58 & 3.41 & 3.25 &2.96 & \textbf{2.29} & 2.96 & 3.32 \\
			&   & $F_\beta \uparrow$     & 61.83 & 77.05 & 52.80 & 70.46 & 80.24 & 82.67 & 82.80 & 85.13          & \textbf{86.35} & 84.75 & 81.88 \\
			&   & $S_m \uparrow$         & 76.46 & 85.65 & 68.74 & 81.35 & 85.48 & 87.20 & 88.19 & 88.67          & \textbf{90.17} & 89.13 & 86.45 \\
			&   & $E_m \uparrow$         & 77.70 & 87.36 & 72.13 & 84.39 & 91.42 & 93.63 & 93.08 & \textbf{94.64} & 94.29          & 94.52 & 92.30 \\
			\cmidrule{2-14}
			& \multirow{4}[2]{*}{COD10K \cite{fan2020camouflaged}} & $MAE\downarrow$   & 8.06 & 5.29 & 10.54 & 5.11 & 4.17 & 4.26 & 3.96 & 3.16 & \textbf{2.89} & 3.53 & 3.62 \\
			&   & $F_\beta \uparrow$     & 48.87 & 59.53 & 41.68 & 61.51 & 65.69 & 67.93 & 70.11 & 75.12          & \textbf{76.56} & 72.05 & 70.93 \\
			&   & $S_m \uparrow$         & 69.62 & 75.01 & 63.43 & 76.68 & 76.97 & 77.64 & 79.98 & 82.23          & \textbf{83.84} & 80.89 & 79.12 \\
			&   & $E_m \uparrow$         & 71.17 & 77.63 & 67.82 & 80.83 & 83.24 & 86.42 & 87.73 & \textbf{89.95} & 88.80          & 88.41 & 86.29 \\
			\cmidrule{2-14}
			& \multirow{4}[2]{*}{NC4K \cite{lv2021simultaneously}} & $MAE\downarrow$   & 8.84 & 7.20 & -- & 6.39 & 5.55 & 5.76 & 5.27 & 4.43 & \textbf{4.34} & 4.65 & 5.05 \\
			&  & $F_\beta \uparrow$     & 63.96 & 70.53 & -- & 72.88 & 76.42 & 76.86 & 78.44 & \textbf{82.41} & 81.75 & 80.62 & 79.88          \\
			&  & $S_m \uparrow$         & 75.75 & 78.74 & -- & 81.08 & 81.22 & 80.80 & 82.90 & 84.70          & 85.28 & 84.15 & \textbf{88.01} \\
			&  & $E_m \uparrow$         & 77.27 & 80.81 & -- & 84.49 & 86.23 & 87.13 & 88.77 & \textbf{90.70 }& 89.60 & 89.84 & 90.23          \\
			\bottomrule
		\end{tabular}%
		\label{tab:supervised}}%
\end{table*}%
\begin{table*}[htbp]
	\centering
	\large
    \setlength{\tabcolsep}{4pt} 
	\caption{Performance ($\%$) of different methods on SOD and COD benchmarks under the unsupervised setting. ``--'' represents the authors have not released the results.}
	\scalebox{0.54}{\begin{tabular}{c|c|c|c|c|c|c|c|c|c|c|c}
					\toprule
					\multicolumn{1}{r}{} & \multicolumn{1}{r}{} & & \multicolumn{5}{c|}{SOD methods} & COD method & Universal method & \multicolumn{2}{c}{Joint learning methods}\\
					\midrule
					\multicolumn{1}{r}{} & \multicolumn{2}{c|}{} & SBF \cite{zhang2017supervision}   & USPS \cite{nguyen2019deepusps}  & UMNet \cite{wang2022multi} & A2S \cite{zhou2022activation}   & A2Sv2 \cite{zhou2023texture} & UCOS \cite{zhang2023unsupervised}  & FOUND \cite{simeoni2023unsupervised} & Ours$_{M-C}$ &Ours$_{D-C}$\\
					\midrule
					\multicolumn{1}{r}{} & \multicolumn{1}{c}{} & & Task-specific & Task-specific & Task-specific & Task-specific & Task-specific & Task-specific & Task-agnostic & \multicolumn{2}{c}{Task-agnostic}  \\
					\midrule

			\multirow{20}[10]{*}{SOD}& \multirow{4}[2]{*}{ECSSD \cite{shi2015hierarchical}} & $MAE\downarrow$   & 8.80 & 6.11 & 6.36 & 6.40 & 4.41 & 4.87 & 5.11 & {4.36} & \textbf{4.22} \\
			&    & $F_\beta \uparrow$     & 79.84 & 87.00 & 87.74 & 88.87 & {91.43} & 88.83 & 89.41 &  91.33     &\textbf{91.80} \\
			&    & $S_m \uparrow$         & 83.23 & 85.66 & 86.77 & 86.65 & 89.35          & 87.83 & 87.53 & {89.82} &\textbf{89.97} \\
			&    & $E_m \uparrow$         & 85.01 & 88.75 & 89.89 & 90.91 & 93.66          & 93.13 & 93.03 & {94.05}  &\textbf{94.23}  \\
			\cmidrule{2-12}
			& \multirow{4}[2]{*}{DUTS \cite{wang2017learning}} & $MAE\downarrow$   & 10.69 & 6.57 & 6.67 & 6.46 & \textbf{4.68}  & -- & 6.07 & 5.32 &4.77 \\
			&    & $F_\beta \uparrow$     & 62.70 & 71.98 & 74.99 & 75.91 & \textbf{81.47} & -- & 74.97 & 78.24 & 80.71 \\
			&    & $S_m \uparrow$         & 68.61 & 77.17 & 80.27 & 81.11 & {84.24} & -- & 80.33 & 83.34 & \textbf{84.55} \\
			&    & $E_m \uparrow$         & 71.54 & 80.11 & 84.48 & 86.51 & {90.19} & -- & 86.60 & 88.69 & \textbf{90.25} \\
			\cmidrule{2-12}
			& \multirow{4}[2]{*}{DUT-O \cite{yang2013saliency}} & $MAE\downarrow$   & 10.76 & 5.70 & 6.31 & 6.88 & \textbf{6.09} & -- & 8.83 & 6.85 &6.65 \\
			&    & $F_\beta \uparrow$     & 61.20 & 72.51 & 73.67 & 72.76 & \textbf{74.98} & -- & 66.30 & 72.01  &73.16 \\
			&    & $S_m \uparrow$         & 74.73 & 79.03 & 80.47 & 79.50 & \textbf{81.22} & -- & 74.69 & 79.76  &80.33\\
			&    & $E_m \uparrow$         & 76.32 & 81.17 & 83.29 & 84.50 & \textbf{86.35} & -- & 80.24 & 84.39  &84.83\\
			\cmidrule{2-12}
			&  \multirow{4}[2]{*}{PASCAL-S \cite{li2014secrets}} & $MAE\downarrow$   & 13.09 & 10.54 & -- & 10.35 & {7.25} & -- & 7.86 & 7.47 &\textbf{6.90} \\
			&   & $F_\beta \uparrow$     & 69.51 & 74.47 & -- & 78.07 & {82.10} & -- & 79.96 & 81.45  &\textbf{82.41}\\
			&   & $S_m \uparrow$         & 75.79 & 76.54 & -- & 78.70 & {83.00} & -- & 81.09 & 82.70  &\textbf{83.50} \\
			&   & $E_m \uparrow$         & 77.77 & 79.47 & -- & 83.69 & {88.72} & -- & 87.45 & 88.12  &\textbf{89.12}\\
			\cmidrule{2-12}
			&  \multirow{4}[2]{*}{HKU-IS \cite{li2015visual}} & $MAE\downarrow$   & 7.53 & 4.21 & 4.12 & 4.20 & 3.65 & 4.09 & 4.20 & {3.34}  &\textbf{3.23} \\
			&   & $F_\beta \uparrow$     & 80.50 & 87.45 & 88.41 & 88.78 & 90.14 & 86.95 & 87.47 & {90.16}  &\textbf{90.57}\\
			&   & $S_m \uparrow$         & 82.91 & 86.68 & 88.65 & 88.23 & 88.99 & 87.13 & 86.93 & {89.71}  &\textbf{89.99}\\
			&   & $E_m \uparrow$         & 89.33 & 90.64 & 92.67 & 93.53 & 94.24 & 93.46 & 93.62 & {94.94}  &\textbf{95.24}\\
			\midrule
			\midrule
			\multirow{16}[8]{*}{COD} & \multirow{4}[2]{*}{CAMO \cite{le2019anabranch}} & $MAE\downarrow$   & -- & -- & -- & 13.43 & 17.32 & 12.70 & 12.89 & {11.94}  &\textbf{11.35}\\
			&   & $F_\beta \uparrow$     & -- & -- & -- & 62.80 & 21.65 & 64.56 & 63.31 & {66.39}  &\textbf{67.83}\\
			&   & $S_m \uparrow$         & -- & -- & -- & 67.08 & 44.57 & 70.04 & 68.54 & {71.57}  &\textbf{72.29}\\
			&   & $E_m \uparrow$         & -- & -- & -- & 74.93 & 38.54 & 78.41 & 78.20 & {78.79}  &\textbf{79.77}\\
			\cmidrule{2-12}
			& \multirow{4}[2]{*}{CHAMELEON \cite{skurowski2018animal}} & $MAE\downarrow$   & -- & -- & -- & 8.78 & 13.42 & 9.53 & 9.51 & {8.34}  &\textbf{8.14} \\
			&   & $F_\beta \uparrow$     & -- & -- & -- & 63.61 & 17.97 & 62.90 & 58.96 & {65.33}  &\textbf{65.39}\\
			&   & $S_m \uparrow$     	 & -- & -- & -- & 70.41 & 45.43 & 71.49 & 68.43 & \textbf{73.20}  &72.91\\
			&   & $E_m \uparrow$     	 & -- & -- & -- & 81.31 & 38.11 & 80.18 & 81.01 & {82.80}  &\textbf{83.32}\\
			\cmidrule{2-12}
			& \multirow{4}[2]{*}{COD10K \cite{fan2020camouflaged}} & $MAE\downarrow$   & -- & -- & -- & 8.55 & 8.45 & 8.62 & 8.50 & {7.83} &\textbf{7.37}\\
			&  & $F_\beta \uparrow$      & -- & -- & -- & 50.53 & 29.22 & 54.61 & 51.95 & {54.84}  &\textbf{56.46}\\
			&  & $S_m \uparrow$     	 & -- & -- & -- & 66.28 & 51.96 & 68.89 & 67.03 & {69.52}  &\textbf{70.36}\\
			&  & $E_m \uparrow$     	 & -- & -- & -- & 73.76 & 52.67 & 73.95 & 75.07 & {76.19}  &\textbf{78.00}\\
			\cmidrule{2-12}
			&  \multirow{4}[2]{*}{NC4K \cite{lv2021simultaneously}} & $MAE\downarrow$   & -- & -- & -- & 9.38 & 13.59 & 8.52 & 8.39 & {7.89} & \textbf{7.57} \\
			&  & $F_\beta \uparrow$      & -- & -- & -- & 65.95 & 36.36 & 68.93 & 67.40 & {70.56}  &\textbf{71.66} \\
			& & $S_m \uparrow$     		 & -- & -- & -- & 72.14 & 51.18 & 75.45 & 74.12 & {76.35}  &\textbf{76.87}\\
			& & $E_m \uparrow$     		 & -- & -- & -- & 80.27 & 48.95 & 81.92 & 82.44 & {83.28}  &\textbf{84.18}\\
			\bottomrule
		\end{tabular}%
		\label{tab:unsupervised}}%
\end{table*}%
\subsubsection{Benchmarks} 
The benchmarks for evaluation includes SOD and COD datasets. 

\textbf{ECSSD} \cite{shi2015hierarchical} contains 1,000 images with complicated structures, which are collected from the Internet.

\textbf{HKU-IS} \cite{li2015visual} consists of 3000 training images and 1,447 test images, which are with multiple disconnected objects.

\textbf{DUTS} \cite{wang2017learning} contains 10,533 training images and 5019 test images, which are with different scenes and various sizes.

\textbf{DUT-OMRON} \cite{yang2013saliency} has 5,168 images with different sizes and complex structures.

In terms of HKU-IS \cite{li2015visual} and DUTS \cite{wang2017learning}, only the test images are used for evaluations in our experiments.

\textbf{CHAMELEON} \cite{skurowski2018animal} is an unpublished dataset that has only 76 images collected from the Internet via the Google search engine using ``camouflaged animal'' as a keyword.

\textbf{CPD1K} \cite{li2014secrets} is the earliest dataset for camouflaged people detection, which contains 1,000 images covering two scene types, namely woodland and snowfield. The test subset has 400 images.

\textbf{COD10K} \cite{fan2020camouflaged}, which is collected from multiple photography websites, contains 10,000 images, including 5,066 camouflaged images, 3,000 background images, and 1,934 non-camouflaged images. The test subset includes 2,026 images.

\textbf{CAMO} \cite{le2019anabranch} has 1,250 images, which are divided into 1,000 training images and 250 testing images.

\textbf{NC4K} \cite{lv2021simultaneously} is a large-scale COD testing dataset, comprising 4,121 images.

\subsubsection{Metrics} 
We evaluate the performance of our model as well as other state-of-the-art methods from both visual and quantitative perspectives. The quantitative metrics include weighted F-measure ($F_\beta$) \cite{achanta2009frequency}, Mean Absolute Error ($MAE$) \cite{achanta2009frequency}, S-measure ($S_m$) \cite{fan2017structure}, and E-measure ($E_m$) \cite{fan2018enhanced}.
Given a continuous saliency map, a binary mask $\hat B$ is achieved by thresholding the saliency map $B$. Precision is defined as $Precision = {{\left| {\hat B \cap G} \right|} \mathord{\left/
		{\vphantom {{\left| {\hat B \cap G} \right|} {\left| \hat B \right|}}} \right.
		\kern-\nulldelimiterspace} {\left| \hat B \right|}}$, and recall is defined as $Recall = {{\left| {\hat B \cap G} \right|} \mathord{\left/
		{\vphantom {{\left| {\hat B \cap G} \right|} {\left| G \right|}}} \right.
		\kern-\nulldelimiterspace} {\left| G \right|}}$. Then, the PR curve is plotted under different thresholds.

F-measure is an overall performance indicator, which is computed by
\begin{equation}
	\label{equ:F-measure}
	{F_\beta } = \frac{{\left( {1 + {\beta ^2}} \right)Precision \times Recall}}{{{\beta ^2}Precision + Recall}}.
\end{equation}
As suggested in \cite{achanta2009frequency}, ${{\beta ^2} = 0.3}$.

$MAE$ is defined as
\begin{equation}
	\label{equ:MAE}
	MAE = \frac{1}{{\hat W \times \hat H}}{\sum\limits_{i} {\left| {B\left( {i} \right) - G\left( {i} \right)} \right|} } ,
\end{equation}
where $\hat W$ and $\hat H$ are the width and height of the image, respectively.

S-measure ($S_m$) \cite{fan2017structure} is computed by
\begin{equation}
	\label{equ:S}
	{S_m} = \alpha {S_o} + \left( {1 - \alpha } \right){S_r},
\end{equation}
where $S_o$ and $S_r$ represent the object-aware and region-aware structure similarities between the prediction and the ground truth, respectively. $\alpha$ is set to 0.5 \cite{fan2017structure}.

E-measure ($E_m$) \cite{fan2018enhanced} combines local pixel values with the image-level mean value to jointly evaluate the similarity between the prediction and the ground truth.

\subsubsection{SOTAs} 
The SOTA methods for comparison include one supervised joint learning method (UJSC \cite{li2021uncertainty}), six supervised SOD methods (PiCANet \cite{liu2018picanet}, CPD \cite{wu2019cascaded}, BASNet \cite{qin2019basnet}, ITSD \cite{zhou2020interactive}, MINet \cite{pang2020multi}, and PFSN \cite{wu2023pixel}), and four supervised COD methods (SINet \cite{fan2020camouflaged}, PFNet \cite{mei2021camouflaged}, FEDER \cite{he2023camouflaged}, and ZoomNet \cite{pang2022zoom}).

\subsubsection{Quantitative comparison} 

Table \ref{tab:supervised} lists the performance of different methods on SOD and COD benchmarks under the supervised setting. To achieve both SOD and COD results, we re-train the SOD/COD methods using COD/SOD training datasets to achieve their COD/SOD results. As listed in Table \ref{tab:supervised}, we conclude five findings: i) SOD methods trained using COD datasets cannot get consistently good performance for COD because camouflaged objects are hard than the salient objects for identification; ii) COD methods trained using SOD datasets can get good SOD performance because salient objects are easy samples for COD; iii) The only joint learning UJSC \cite{li2021uncertainty} gets consistently good performance for both SOD and COD, but its task-specificity cannot solve the unforewarnable case; iv) As the only task-agnostic method, our method performs well for SOD and COD, and competitively with the previous task-specific methods.

It is reasonable that our model is slightly inferior to UJSC \cite{li2021uncertainty} and ZoomNet \cite{pang2022zoom} because they are task-specific. More specifically, ZoomNet \cite{pang2022zoom} is trained using COD datasets. UJSC \cite{li2021uncertainty} is trained using SOD and COD datasets for SOD decoder and COD decoder separately. However, our model is trained using the joint SOD and COD datasets, which is task-agnostic but introduces some domain bias.

\subsubsection{Visual comparison}
\begin{figure*}[ht]
	\centering
	\includegraphics[width=0.90\linewidth]{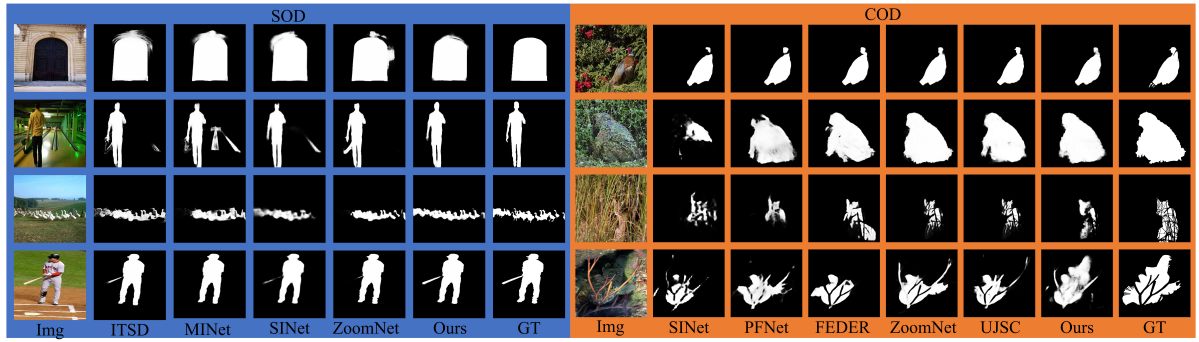}	
	\caption{Visual comparison for SOD and COD in the supervised setting.}
	\label{fig:supervised}
\end{figure*}

For better displaying the joint learning of SOD and COD, Figs. \ref{fig:supervised} shows visual detections for SOD and COD in the supervised setting, respectively. As displayed in Fig. \ref{fig:supervised}, using the unified framework, we get detection results more closer to ground truth in terms of object wholeness and uniformity in the supervised circumstance. 
\subsection{Supervised learning}

\subsection{Unsupervised learning}

\subsubsection{SOTAs} The chosen SOTA methods for comparison include five unsupervised SOD methods (SBF \cite{zhang2017supervision}, USPS \cite{nguyen2019deepusps}, UMNet \cite{wang2022multi}, A2S \cite{zhou2022activation}, and A2S v2 \cite{zhou2023texture}), (only) one unsupervised COD method (UCOS \cite{zhang2023unsupervised}), and one unsupervised universal image segmentation method (FOUND \cite{simeoni2023unsupervised}).
\begin{figure*}[ht]
	\centering
	\includegraphics[width=0.90\linewidth]{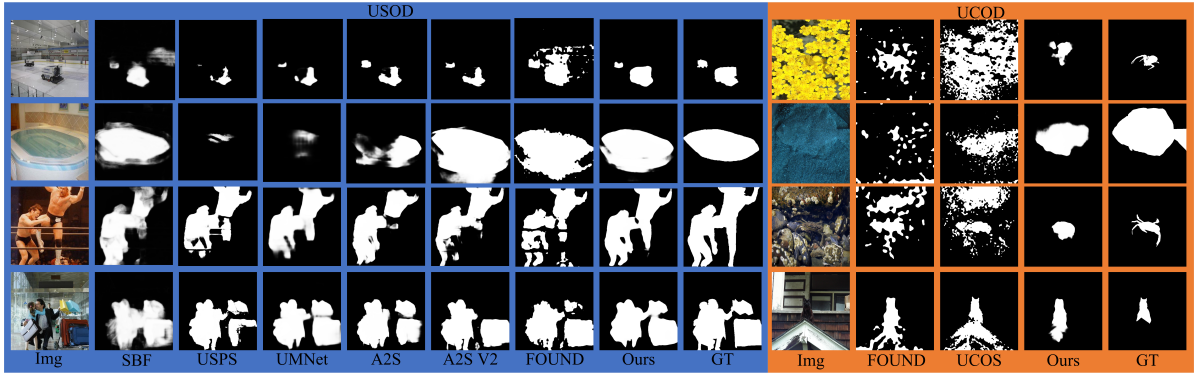}
	
	\caption{Visual comparison for SOD and COD in the unsupervised setting.}
	\label{fig:unsupervised}
\end{figure*}
\begin{table*}[htbp]
	\centering
    \footnotesize
	\caption{Ablation study ($\%$) for CDP in the supervised setting. ``$\ast-$CDP'' is achieved by replacing IGC with ``$\ast$'' model in our framework.}
    \setlength{\tabcolsep}{3pt} 
    \renewcommand{\arraystretch}{0.8}  
	\scalebox{0.84}{\begin{tabular}{c|c|c|c|c|c|c|c|c|c|c}
			\toprule
			\multicolumn{3}{c|}{} & CPD \cite{wu2019cascaded}   & CPD-CDP & \multirow{37}[20]{*}{} & ITSD \cite{zhou2020interactive}  & ITSD-CDP & \multirow{37}[20]{*}{} & MINet \cite{pang2020multi} & MINet-CDP \\
			\cmidrule{1-5}\cmidrule{7-8}\cmidrule{10-11}    \multicolumn{1}{c|}{\multirow{20}[10]{*}{SOD}} & \multirow{4}[2]{*}{ECSSD} & $MAE \downarrow$   & \textbf{4.02} & 4.60  &       & 4.01  & \textbf{3.65 } &       & \textbf{3.62} & 3.95 \\
			&       & $F_\beta \uparrow$     & \textbf{91.15} & 89.36  &       & 91.01 & \textbf{91.59 } &       & \textbf{91.87} & 91.32 \\
			&       & $S_m \uparrow$     & \textbf{91.02} & 89.62  &       & 91.42 & \textbf{91.43 } &       & \textbf{91.91} & 91.07 \\
			&       & $E_m \uparrow$     & \textbf{93.77} & 91.76  &       & 93.75 & \textbf{94.01 } &       & \textbf{94.32} & 93.45 \\
			\cmidrule{2-5}\cmidrule{7-8}\cmidrule{10-11}          & \multirow{4}[2]{*}{DUTS} & $MAE \downarrow$   & 4.29  & \textbf{3.19} &       & 4.23  & \textbf{3.24 } &       & 3.94  & \textbf{3.4} \\
			&       & $F_\beta \uparrow$     & 82.44 & \textbf{86.15} &       & 83.23 & \textbf{86.27 } &       & 83.49 & \textbf{85.89} \\
			&       & $S_m \uparrow$     & 86.66 & \textbf{89.26} &       & 87.71 & \textbf{89.30 } &       & 87.49 & \textbf{88.82} \\
			&       & $E_m \uparrow$     & 90.2  & \textbf{93.21} &       & 90.56 & \textbf{93.29 } &       & 90.67 & \textbf{92.54} \\
			\cmidrule{2-5}\cmidrule{7-8}\cmidrule{10-11}          & \multirow{4}[2]{*}{DUT-O} & $MAE \downarrow$   & 5.67  & \textbf{4.91} &       & 6.32  & \textbf{4.93 } &       & 5.69  & \textbf{5.43} \\
			&       & $F_\beta \uparrow$     & 73.85 & \textbf{76.75} &       & 75.24 & \textbf{77.96 } &       & 74.04 & \textbf{75.48} \\
			&       & $S_m \uparrow$     & 81.77 & \textbf{83.79} &       & 82.88 & \textbf{84.47 } &       & 82.18 & \textbf{82.82} \\
			&       & $E_m \uparrow$     & 84.5  & \textbf{86.88} &       & 85.28 & \textbf{88.11 } &       & 84.58 & \textbf{86.16} \\
			\cmidrule{2-5}\cmidrule{7-8}\cmidrule{10-11}          & \multirow{4}[2]{*}{PASCAL-S} & $MAE \downarrow$   & 7.21  & \textbf{6.36} &       & 6.81  & \textbf{5.97 } &       & \textbf{6.39} & 6.47  \\
			&       & $F_\beta \uparrow$     & 82.3  & \textbf{83.38} &       & 83.05 & \textbf{84.09 } &       & 83.03 & \textbf{83.19 } \\
			&       & $S_m \uparrow$     & 84.46 & \textbf{85.3} &       & 85.63 & \textbf{85.78}  &       & \textbf{85.54} & 85.21  \\
			&       & $E_m \uparrow$     & 88.25 & \textbf{88.53} &       & 89.15 & \textbf{89.56 } &       & \textbf{89.36} & 88.67  \\
			\cmidrule{2-5}\cmidrule{7-8}\cmidrule{10-11}          & \multirow{4}[2]{*}{HKU-IS} & $MAE \uparrow$   & \textbf{3.32} & 3.57  &       & 3.46  & \textbf{3.15 } &       & \textbf{3.03} & 3.19 \\
			&       & $F_\beta \uparrow$     & \textbf{89.58} & 88.88 &       & 89.4  & \textbf{89.98 } &       & \textbf{90.55} & 90.14 \\
			&       & $S_m \uparrow$     & \textbf{90.45} & 89.75 &       & \textbf{90.68} & 90.52  &       & \textbf{91.39} & 90.6 \\
			&       & $E_m \uparrow$     & \textbf{94.24} & 92.88 &       & \textbf{93.95} & 93.90  &       & \textbf{94.65} & 93.91 \\
			\cmidrule{1-5}\cmidrule{7-8}\cmidrule{10-11}    \multicolumn{1}{c|}{\multirow{16}[8]{*}{COD}} & \multirow{4}[2]{*}{CAMO} & $MAE \downarrow$   & 11.29 & \textbf{7.92} &       & 10.16 & \textbf{8.33 } &       & 9.03  & \textbf{8.8} \\
			&       & $F_\beta \uparrow$     & 61.77 & \textbf{75.37} &       & 66.29 & \textbf{73.90 } &       & 69.12 & \textbf{71.97} \\
			&       & $S_m \uparrow$     & 71.61 & \textbf{79.43} &       & 74.99 & \textbf{77.83 } &       & 74.8  & \textbf{76.56} \\
			&       & $E_m \uparrow$     & 72.27 & \textbf{84.94} &       & 77.99 & \textbf{83.87 } &       & 79.18 & \textbf{82.35} \\
			\cmidrule{2-5}\cmidrule{7-8}\cmidrule{10-11}          & \multirow{4}[2]{*}{CHAMELEON} & $MAE \downarrow$   & 4.8   & \textbf{3.26} &       & 5.73  & \textbf{3.41 } &       & \textbf{3.58}  & 3.65 \\
			&       & $F_\beta \uparrow$     & 77.05 & \textbf{81.65} &       & 70.46 & \textbf{81.52 } &       & \textbf{80.24} & 79.71 \\
			&       & $S_m \uparrow$     & 85.65 & \textbf{86.81} &       & 81.35 & \textbf{86.14 } &       & \textbf{85.48 } & 84.67\\
			&       & $E_m \uparrow$     & 87.36 & \textbf{93.06} &       & 84.39 & \textbf{92.69 } &       & \textbf{91.42} & 91.08  \\
			\cmidrule{2-5}\cmidrule{7-8}\cmidrule{10-11}          & \multirow{4}[2]{*}{COD10K} & $MAE \downarrow$   & 5.29  & \textbf{3.67} &       & 5.11  & \textbf{4.01 } &       & 4.17  & \textbf{4.01} \\
			&       & $F_\beta \uparrow$     & 59.53 & \textbf{70.57} &       & 61.51 & \textbf{69.01 } &       & 65.69 & \textbf{69.84} \\
			&       & $S_m \uparrow$     & 75.01 & \textbf{79.67} &       & 76.68 & \textbf{78.51 } &       & 76.97 & \textbf{78.92} \\
			&       & $E_m \uparrow$     & 77.63 & \textbf{87.22} &       & 80.83 & \textbf{86.43 } &       & 83.24 & \textbf{86.21} \\
			\cmidrule{2-5}\cmidrule{7-8}\cmidrule{10-11}          & \multirow{4}[2]{*}{NC4K} & $MAE \downarrow$   & 7.2   & \textbf{4.82} &       & 6.39  & \textbf{5.10 } &       & 5.55  & \textbf{5.12} \\
			&       & $F_\beta \uparrow$     & 70.53 & \textbf{79.71} &       & 72.88 & \textbf{78.78 } &       & 76.42 & \textbf{78.94} \\
			&       & $S_m \uparrow$     & 78.74 & \textbf{83.65} &       & 81.08 & \textbf{82.94 } &       & 81.22 & \textbf{82.86} \\
			&       & $E_m \uparrow$     & 80.81 & \textbf{88.96} &       & 84.49 & \textbf{88.51 } &       & 86.23 & \textbf{88.17} \\
			\bottomrule
		\end{tabular}%
		\label{tab:plug}}%
\end{table*}%
\begin{table*}[htbp]
	\centering
	
	\caption{Ablation study ($\%$) of CDP for SOD in the supervised setting.}
    \setlength{\tabcolsep}{2pt} 
	\scalebox{0.72}{
		\begin{tabular}{c|cccc|cccc|cccc|cccc|cccc}
			\toprule
			\multirow{2}[2]{*}{method} & \multicolumn{4}{c|}{ECSSD \cite{shi2015hierarchical} }      & \multicolumn{4}{c|}{DUTS \cite{wang2017learning} }      & \multicolumn{4}{c|}{DUT-O \cite{yang2013saliency} }     & \multicolumn{4}{c|}{PASCAL-S \cite{li2014secrets} }  & \multicolumn{4}{c}{HKU-IS \cite{li2015visual} } \\
			& $MAE \downarrow$   & $F_\beta \uparrow$     & $S_m \uparrow$     & $E_m \uparrow$     & $MAE \downarrow$   & $F_\beta \uparrow$     & $S_m \uparrow$     & $E_m \uparrow$     & $MAE \downarrow$   & $F_\beta \uparrow$     & $S_m \uparrow$     & $E_m \uparrow$     & $MAE \downarrow$   & $F_\beta \uparrow$     & $S_m \uparrow$     & $E_m \uparrow$     & $MAE \downarrow$   & $F_\beta \uparrow$     & $S_m \uparrow$     & $E_m \uparrow$ \\
			\midrule
			CPD$_D$ \cite{wu2019cascaded}   & \textbf{4.02 } & \textbf{91.15 } & \textbf{91.02 } & \textbf{93.77 } & 4.29  & 82.44  & 86.66  & 90.20  & 5.67  & 73.85  & 81.77  & 84.50  & 7.21  & 82.30  & 84.46  & 88.25  & \textbf{3.32 } & \textbf{89.58 } & 90.45  & \textbf{94.24 } \\
			CPD$_{D-C}$ & 4.98  & 88.97  & 90.59  & 91.44  & 4.53  & 81.15  & 87.38  & 88.83  & 5.91  & 72.66  & 82.55  & 83.66  & 9.12  & 77.30  & 82.03  & 82.57  & 3.75  & 88.11  & \textbf{90.80 } & 92.69  \\
			CPD-CDP$_{D-C}$ & 4.60  & 89.36  & 89.62  & 91.76  & \textbf{3.19 } & \textbf{86.15 } & \textbf{89.26 } & \textbf{93.21 } & \textbf{4.91 } & \textbf{76.75 } & \textbf{83.79 } & \textbf{86.88 } & \textbf{6.36 } & \textbf{83.38 } & \textbf{85.30 } & \textbf{88.53 } & 3.57  & 88.88  & 89.75  & 92.88  \\
			\midrule
			SINet$_D$ \cite{fan2020camouflaged} & \textbf{3.58 } & \textbf{91.75 } & \textbf{92.39 } & \textbf{94.61 } & 4.10  & 82.83  & 87.87  & 91.78  & 5.63  & 75.16  & \textbf{83.24 } & 85.81  & \textbf{6.65 } & \textbf{82.78 } & \textbf{85.89 } & \textbf{89.22 } & \textbf{3.21 } & \textbf{89.82 } & \textbf{91.44 } & \textbf{94.53 } \\
			SINet$_{D-C} $& 4.46  & 90.23  & 91.00  & 92.69  & 3.98  & 83.44  & \textbf{88.31 } & 90.77  & 5.54  & 74.35  & 82.99  & 85.03  & 7.45  & 81.72  & 84.74  & 86.83  & 3.29  & 89.61  & 91.38  & 94.00  \\
			SINet-CDP$_{D-C}$ & 3.91  & 91.20  & 91.26  & 93.73  & \textbf{3.66 } & \textbf{84.75 } & 88.06  & \textbf{92.22 } & \textbf{5.12 } & \textbf{76.15 } & \textbf{83.24 } & \textbf{86.72 } & 6.87  & 82.68  & 84.71  & 88.36  & 3.33  & 89.64  & 90.31  & 93.86  \\
			\bottomrule
	\end{tabular}}%
	\label{tab:cdp_sod}%
\end{table*}%
\begin{table*}[htbp]
	\centering
    \footnotesize 
	\caption{Ablation study ($\%$) of CDP for COD in the supervised setting.}
    \setlength{\tabcolsep}{2pt} 
	\scalebox{0.87}{
		\begin{tabular}{c|cccc|cccc|cccc|cccc}
			\toprule
			\multirow{2}[2]{*}{method} & \multicolumn{4}{c|}{CAMO \cite{le2019anabranch} }      & \multicolumn{4}{c|}{CHAMELEON \cite{skurowski2018animal} } & \multicolumn{4}{c|}{COD10K \cite{fan2020camouflaged} }    & \multicolumn{4}{c}{NC4K \cite{lv2021simultaneously}  } \\
			& $MAE \downarrow$   & $F_\beta \uparrow$     & $S_m \uparrow$     & $E_m \uparrow$     & $MAE \downarrow$   & $F_\beta \uparrow$     & $S_m \uparrow$     & $E_m \uparrow$     & $MAE \downarrow$   & $F_\beta \uparrow$     & $S_m \uparrow$     & $E_m \uparrow$     &$MAE \downarrow$   & $F_\beta \uparrow$     & $S_m \uparrow$     & $E_m \uparrow$ \\
			\midrule
			CPD$_C$ \cite{wu2019cascaded}  & 11.29  & 61.77  & 71.61  & 72.27  & 4.80  & 77.05  & 85.65  & 87.36  & 5.29  & 59.53  & 75.01  & 77.63  & 7.20  & 70.53  & 78.74  & 80.81  \\
			CPD$_{D-C}$ & 11.39  & 61.15  & 71.06  & 72.69  & 5.16  & 74.99  & 84.57  & 85.29  & 5.60  & 59.10  & 75.38  & 76.68  & 5.60  & 70.90  & 79.68  & 81.09  \\
			CPD-CDP$_{D-C}$ & \textbf{7.92 } & \textbf{75.37 } & \textbf{79.43 } & \textbf{84.94 } & \textbf{3.26 } & \textbf{81.65 } & \textbf{86.81 } & \textbf{93.06 } & \textbf{3.67 } & \textbf{70.57 } & \textbf{79.67 } & \textbf{87.22 } & \textbf{4.82 } & \textbf{79.70 } & \textbf{83.65 } & \textbf{88.96 } \\
			\midrule
			SINet$_C$ \cite{fan2020camouflaged} & \textbf{9.15 } & \textbf{70.20 } & \textbf{74.54 } & \textbf{80.35 } & \textbf{3.41 } & \textbf{82.67 } & 87.20  & \textbf{93.63 } & \textbf{4.26 } & \textbf{67.93 } & \textbf{77.64 } & \textbf{86.42 } & 5.76  & 76.86  & 80.80  & 87.13  \\
			SINet$_{D-C}$ & 10.10  & 66.58  & 74.51  & 76.54  & 3.75  & 80.50  & \textbf{88.12}  & 89.63  & 4.79  & 64.07  & 77.41  & 80.50  & 6.13  & 74.75  & 81.45  & 84.18  \\
			SINet-CDP$_{D-C}$ & 9.62  & 68.23  & 74.50  & 79.72  & 3.49  & 80.41  & 85.96  & 92.69  & 4.40  & 66.72  & 77.35  & 84.72  & \textbf{5.58 } & \textbf{77.19 } & \textbf{81.94 } & \textbf{87.23 } \\
			\bottomrule
	\end{tabular}}%
	\label{tab:cdp_cod}%
\end{table*}%
\begin{table*}[htbp]
	\centering
	\caption{Ablation study ($\%$) in the unsupervised setting.}
    
	\large 
	\scalebox{0.50}{\begin{tabular}{c|c|c|c|c|c|c|c|c|c|c|c|c|c|c|c|c}
			\toprule
			& \multirow{2}[4]{*}{} & \multirow{2}[4]{*}{} & \multicolumn{3}{c|}{(a) Different components} & \multirow{38}[22]{*}{} & \multicolumn{4}{c|}{(b) Different initial pseudo masks} & \multirow{38}[22]{*}{} & \multicolumn{5}{c}{(c) Different background semantics} \\
			\cmidrule{4-6}\cmidrule{8-11}\cmidrule{13-17}          &       &       & BASE & BASE+IGC & BASE+IGC+CDP &       & TokenCut \cite{wang2022self} & SpectralSeg \cite{melas2022deep} & SelfMask \cite{shin2022unsupervised}& Ours &       & $\textbf{E}_0$    & $\textbf{E}_1$    & $\textbf{E}_2$    & $\textbf{E}_3$    & $\textbf{E}_4$ \\
			
			\cmidrule{1-6}\cmidrule{8-11}\cmidrule{13-17}    \multirow{20}[10]{*}{SOD} & \multirow{4}[2]{*}{ECSSD \cite{shi2015hierarchical}} & $MAE\downarrow$   & 5.09 & 4.63 & \textbf{4.22} & & 8.55 & 14.21 & 5.89 & \textbf{5.09} & & {4.35} & 4.38 & \textbf{4.22} & 4.36 & 4.38 \\
			&       & $F_\beta\uparrow$ & 89.23 & 90.71 & \textbf{91.80} &       & 84.84 & 69.80 & 89.19 & \textbf{89.23} &       & 91.27 & 91.26 & \textbf{91.80} & 91.27 & 91.18 \\
			&       & $S_m\uparrow$     & 87.36 & 88.73 & \textbf{89.97} &       & 82.77 & 75.98 & 86.24 & \textbf{87.36} &       & 89.75 & 89.76 & \textbf{89.97} & 89.78 & 89.63 \\
			&       & $E_m\uparrow$     & 93.01 & 92.87 & \textbf{94.23} &       & 86.97 & 80.37 & 91.90 & \textbf{93.01} &       & 94.03 & 93.99 & \textbf{94.23} & 94.01 & 93.85 \\
			
			\cmidrule{2-6}\cmidrule{8-11}\cmidrule{13-17}          & \multirow{4}[2]{*}{DUTS \cite{wang2017learning}} & $MAE\downarrow$   & 6.22 & 5.53 & \textbf{4.77} & & 8.87 & 16.42 & \textbf{5.58} & 6.22 & & 5.48 & 5.39 & \textbf{4.77} & 5.46 & 5.40 \\
			&       & $F_\beta\uparrow$ & 73.87 & 77.85 & \textbf{80.71} &       & 74.17 & 52.53 & \textbf{78.69} & 73.87 &       & 77.83 & 78.23 & \textbf{80.71} & 77.91 & 77.95 \\
			&       & $S_m\uparrow$     & 79.85 & 82.41 & \textbf{84.55} &       & 77.18 & 66.20 & \textbf{81.36} & 79.85 &       & 83.25 & 83.32 & \textbf{84.55} & 83.19 & 83.30 \\
			&       & $E_m\uparrow$     & 85.90 & 87.54 & \textbf{90.25} &       & 82.67 & 69.01 & \textbf{88.63} & 85.90 &       & 88.27 & 88.55 & \textbf{90.25} & 88.33 & 88.48 \\
			
			\cmidrule{2-6}\cmidrule{8-11}\cmidrule{13-17}          & \multirow{4}[2]{*}{DUT-O \cite{yang2013saliency}} & $MAE\downarrow$   & 9.14 & 7.14 & \textbf{6.65} & & 10.61 & 19.67 & \textbf{6.55} & 9.14 & & 7.29 & 7.14 & \textbf{6.65} & 7.23 & 7.18 \\
			&       & $F_\beta\uparrow$ & 65.14 & 71.41 & \textbf{73.16} &       & 68.35 & 47.16 & \textbf{73.12} & 65.14 &       & 71.04 & 71.40 & \textbf{72.01} & 71.15 & 71.11 \\
			&       & $S_m\uparrow$     & 74.11 & 78.61 & \textbf{80.33} &       & 75.21 & 62.47 & \textbf{72.01} & 74.11 &       & 79.37 & 79.48 & \textbf{79.76} & 79.32 & 79.27 \\
			&       & $E_m\uparrow$     & 79.39 & 83.76 & \textbf{84.83} &       & 80.33 & 64.29 & \textbf{85.78} & 79.39 &       & 83.39 & 83.73 & \textbf{84.83} & 83.50 & 83.21 \\
			
			\cmidrule{2-6}\cmidrule{8-11}\cmidrule{13-17}          & \multirow{4}[2]{*}{PASCAL-S \cite{li2014secrets}} & $MAE\downarrow$   & 7.76 & 7.83 & \textbf{6.90} & & 12.44 & 19.18 & 8.44 & \textbf{7.76} & & 7.52 & 7.51 & \textbf{6.90} & 7.48 & 7.49 \\
			&       & $F_\beta\uparrow$ & 79.64 & 80.86 & \textbf{82.41} &       & 75.86 & 60.28 & \textbf{81.10} & 79.64          &       & 81.08 & 81.24          & \textbf{82.41} & 81.19          & 81.05 \\
			&       & $S_m\uparrow$     & 81.09 & 81.92 & \textbf{83.50} &       & 76.09 & 67.40 & 80.44          & \textbf{81.09} &       & 82.69 & {82.74} & \textbf{83.50}          & 82.69          & 82.60 \\
			&       & $E_m\uparrow$     & 87.41 & 87.33 & \textbf{89.12} &       & 82.22 & 72.15 & 86.75          & \textbf{87.41} &       & 88.05 & 88.07          & \textbf{89.12} & \textbf{88.12} & 88.09 \\
			
			\cmidrule{2-6}\cmidrule{8-11}\cmidrule{13-17}          & \multirow{4}[2]{*}{HKU-IS \cite{li2015visual}} & $MAE\downarrow$  & 4.27  & 3.54 & \textbf{3.23} & & 7.02 & 11.35 & 5.06 & \textbf{4.27} & & {3.34} & 3.35 & \textbf{3.23} & 3.35 & 3.36 \\
			&       & $F_\beta\uparrow$ & 87.69 & 89.53 & \textbf{90.57} &       & 81.67 & 67.44 & 86.80 & \textbf{87.69} &       & 89.69 & 89.78 & \textbf{90.57} & 89.80 & 89.50 \\
			&       & $S_m\uparrow$     & 86.57 & 88.27 & \textbf{89.99} &       & 78.66 & 76.02 & 84.92 & \textbf{86.57} &       & 89.63 & 89.68 & \textbf{89.99} & 89.65 & 89.62 \\
			&       & $E_m\uparrow$     & 93.44 & 93.67 & \textbf{95.24} &       & 84.02 & 80.80 & 92.35 & \textbf{93.44} &       & 94.84 & 94.90 & \textbf{95.24} & 94.85 & 94.84 \\
			
			\cmidrule{1-6}\cmidrule{8-11}\cmidrule{13-17} \midrule  \multirow{16}[8]{*}{COD} & \multirow{4}[2]{*}{CAMO \cite{le2019anabranch}} & $MAE\downarrow$   & 12.74 & 12.35 & \textbf{11.35} & & 16.26 & 23.50 & 18.77 & \textbf{12.74} & & 12.29 & 12.24 & \textbf{11.35} & 12.07 & 12.23 \\
			&       & $F_\beta\uparrow$ & 63.27 & 64.35 & \textbf{67.83} &       & 54.33 & 48.05 & 53.62 & \textbf{63.27} &       & 66.32 & 66.34 & \textbf{67.83} & 66.28 & 66.22 \\
			&       & $S_m\uparrow$     & 67.91 & 70.37 & \textbf{72.29} &       & 63.52 & 57.91 & 61.73 & \textbf{67.91} &       & 71.50 & 71.56 & \textbf{72.29} & 71.50 & 71.44 \\
			&       & $E_m\uparrow$     & 78.28 & 78.14 & \textbf{79.77} &       & 70.63 & 64.76 & 69.79 & \textbf{78.28} &       & 78.35 & 78.34 & \textbf{79.77} & 78.45 & 78.37 \\
			
			\cmidrule{2-6}\cmidrule{8-11}\cmidrule{13-17}          & \multirow{4}[2]{*}{CHAMELEON \cite{skurowski2018animal}} & $MAE\downarrow$   & 9.38 & 8.55 & \textbf{8.14} & & 13.18 & 21.95 & 17.59 & \textbf{9.38} & & 8.43 & 8.38 & \textbf{8.14} & 8.36 & 8.46 \\
			&       & $F_\beta\uparrow$ & 59.02 & 64.26 & \textbf{65.39} &       & 53.56 & 43.96 & 48.09 & \textbf{59.02} &       & 64.90 & 64.99          & \textbf{65.39} & 65.21 & 64.46 \\
			&       & $S_m\uparrow$     & 68.17 & 72.19 & \textbf{72.91} &       & 65.35 & 57.47 & 61.89 & \textbf{68.17} &       & 73.10 & 72.98          & \textbf{72.91} & 73.08 & 72.83 \\
			&       & $E_m\uparrow$     & 80.62 & 82.08 & \textbf{83.32} &       & 73.96 & 62.79 & 67.51 & \textbf{80.62} &       & 82.19 & {83.04} & \textbf{83.32}         & 82.35 & 81.99 \\
			
			\cmidrule{2-6}\cmidrule{8-11}\cmidrule{13-17}          & \multirow{4}[2]{*}{COD10K \cite{fan2020camouflaged}} & $MAE\downarrow$   & 8.75 & 7.98 & \textbf{7.37} & & 10.34 & 19.32 & 13.09 & \textbf{8.75} & & 7.98 & 7.85 & \textbf{7.37} & 7.84 & 7.90 \\
			&       & $F_\beta\uparrow$ & 51.93 & 53.94          & \textbf{56.46} &       & 50.24 & 38.81 & 46.92 & \textbf{51.93} &       & 54.56 & 54.78 & \textbf{56.46} & 54.76 & 54.65 \\
			&       & $S_m\uparrow$     & 66.87 & {69.77} & \textbf{70.36}   &       & 65.78 & 57.51 & 63.71 & \textbf{66.87} &       & 69.48 & 69.56 & \textbf{70.36}         & 69.57 & {69.58} \\
			&       & $E_m\uparrow$     & 76.14 & 75.90          & \textbf{78.00} &       & 73.50 & 59.52 & 67.87 & \textbf{76.14} &       & 75.90 & 76.13 & \textbf{78.00} & 76.17 & 76.09 \\
			
			\cmidrule{2-6}\cmidrule{8-11}\cmidrule{13-17}          & \multirow{4}[2]{*}{NC4K \cite{lv2021simultaneously}} & $MAE\downarrow$   & 8.59 & 8.13 & \textbf{7.57} & & 10.12 & 15.88 & 11.41 & \textbf{7.59} & & 7.98 & 7.94 & \textbf{7.57} & 7.90 & 7.93 \\
			&       & $F_\beta\uparrow$ & 67.35 & 69.92 & \textbf{71.66} &       & 64.88 & 56.18 & 63.38 & \textbf{67.35} &       & 70.34          & 70.44 & \textbf{71.66} & {70.56} & 70.38 \\
			&       & $S_m\uparrow$     & 74.21 & 75.47 & \textbf{76.87} &       & 72.45 & 66.91 & 71.57 & \textbf{74.21} &       & {76.50} & 75.79 & \textbf{76.87}         & 76.32          & 76.39 \\
			&       & $E_m\uparrow$     & 83.04 & 82.76 & \textbf{84.18} &       & 80.21 & 71.90 & 77.65 & \textbf{83.04} &       & 83.18          & 83.05 & \textbf{84.18} & 83.27          & 83.22 \\
			\bottomrule
		\end{tabular}%
		\label{tab:ablation}}
	
\end{table*}%

\subsubsection{Quantitative comparison} 

Table \ref{tab:unsupervised} lists the performance of different unsupervised methods on SOD and COD benchmarks. Similar to the supervised setting, we re-train the unsupervised SOD/COD models using the COD/SOD datasets to achieve COD/SOD results. As shown in Table \ref{tab:unsupervised}, we conclude three findings: i) The unsupervised SOD models trained using the COD datasets cannot well solve the COD task; ii) The unsupervised COD method trained using the SOD datasets cannot well solve the SOD task either; iii) Our task-agnostic model performs better than the previous task-specific SOD and COD methods on most metrics; iv) Our model beats the previous universal FOUND \cite{simeoni2023unsupervised} by a large margin.
\subsubsection{Visual comparison}
\begin{table*}[htbp]
	\centering
	\caption{Improvement for the update of pseudo labels.}
	\normalsize
    \setlength{\tabcolsep}{4pt}  

	\scalebox{0.64}{
		\begin{tabular}{cc|cccc|cccc|cccc|cccc}
			\toprule
			\multicolumn{2}{c}{\multirow{2}[2]{*}{}} & \multicolumn{4}{|c|}{ECSSD \cite{shi2015hierarchical}}     & \multicolumn{4}{c|}{HKU-IS \cite{li2015visual} }    & \multicolumn{4}{c|}{CAMO \cite{le2019anabranch}}      & \multicolumn{4}{c}{COD10K \cite{fan2020camouflaged}} \\
			\multicolumn{2}{c|}{} & $MAE \downarrow$   & $F_\beta \uparrow$     & $S_m \uparrow$     & $E_m \uparrow$     & $MAE \downarrow$   & $F_\beta \uparrow$     & $S_m \uparrow$     & $E_m \uparrow$     & $MAE \downarrow$   & $F_\beta \uparrow$     & $S_m \uparrow$     & $E_m \uparrow$     & $MAE \downarrow$   & $F_\beta \uparrow$     & $S_m \uparrow$     & $E_m \uparrow$ \\
			\midrule
			\multirow{2}[1]{*}{(a)} &Ours(w/o update) & 4.94  & 90.82  & 88.52  & 92.21  & 4.22  & 88.47  & 88.73  & 93.48  & 12.28  & 63.92  & 70.92  & 77.53  & 8.04  & 52.37  & 68.93  & {76.32 } \\
			& Ours  & \textbf{4.22}  & \textbf{91.80}  & \textbf{89.87}  & \textbf{94.23}  & \textbf{3.23}  & \textbf{90.57}  & \textbf{89.99}  & \textbf{95.24}  & \textbf{11.35}  & \textbf{67.83}  &\textbf{72.29}  & \textbf{79.77}  & \textbf{7.37}  & \textbf{56.46}  & \textbf{70.36}  & \textbf{78.00}  \\
			\midrule
			\midrule
			\multirow{3}[1]{*}{(b)} &CPD \cite{wu2019cascaded} & 5.04  & 90.12  & {89.96 } & 92.77  & 3.92  & 88.89  & {89.82 } & 93.58  & 12.97  & 63.13  & 70.92  & 76.23  & 8.28  & 53.52  & {69.93 } & 75.32  \\
			&ZoomNet \cite{pang2022zoom} & 4.83  & 90.91  & 88.89  & 92.93  & 3.56  & 89.72  & 89.23  & 94.33  & 12.56  & 65.73  & 71.08  & 77.45  & 7.90  & {55.79 } & 68.87  & {77.63 } \\
			& Ours  & \textbf{4.22}  & \textbf{91.80}  & \textbf{89.87}  & \textbf{94.23}  & \textbf{3.23}  & \textbf{90.57}  & \textbf{89.99}  & \textbf{95.24}  & \textbf{11.35}  & \textbf{67.83}  &\textbf{72.29}  & \textbf{79.77}  & \textbf{7.37}  & \textbf{56.46}  & \textbf{70.36}  & \textbf{78.00}  \\
			\midrule
			\midrule
			\multirow{3}[1]{*}{(c)} & Ours\_sod & \textbf{3.16 } & \textbf{93.04 } & \textbf{92.82 } & \textbf{95.20 } & \textbf{2.52 } & 91.67  & \textbf{92.25 } & 95.46  & --    & --    & --    & --    & --    & --    & --    & -- \\
			& Ours\_cod & --    & --    & --    & --    & --    & --    & --    & --    & 7.37  & 75.35  & 79.25  & 84.21  & \textbf{3.60 } & 70.82  & \textbf{79.81 } & \textbf{86.78 } \\
			& Ours  & 3.39  & 92.59  & 92.24  & 94.81  & 2.59  & \textbf{91.87 } & 92.16  & \textbf{95.58 } & \textbf{7.17 } & \textbf{76.70 } & \textbf{79.78 } & \textbf{84.44 } & 3.62  & \textbf{70.93 } & 79.12  & 86.29  \\
			\midrule
			\midrule
			\multirow{3}[1]{*}{(d)} & Ours\_sod & \textbf{4.02 } & \textbf{92.35 } & \textbf{90.29 } & \textbf{94.35 } & \textbf{3.14 } & \textbf{91.54 } & \textbf{90.18 } & {94.99 } & --    & --    & --    & --    & --    & --    & --    & -- \\
			& Ours\_cod & --    & --    & --    & --    & --    & --    & --    & --    & {11.42 } & {67.29 } & {71.73 } & {79.38 } & {7.44 } & {55.56 } & {69.74 } & {77.12 } \\
			& Ours  & 4.22  & 91.80  & 89.87  & 94.23  & 3.23  & 90.57  & 89.99  & \textbf{95.24}  & \textbf{11.35}  & \textbf{67.83}  & \textbf{72.29}  & \textbf{79.77}  & \textbf{7.37}  & \textbf{56.46 } & \textbf{70.36}  & \textbf{78.00}  \\
			\bottomrule
	\end{tabular}}%
	\label{tab:ablationv}%
\end{table*}%
For better displaying the joint learning of SOD and COD, Fig. \ref{fig:unsupervised} shows visual detections for SOD and COD in the unsupervised setting, respectively. As displayed in Fig. \ref{fig:unsupervised} in the unsupervised setting, the previous methods mostly introduce lots of noise in the detected saliency map, and even cannot identify the camouflaged objects. Differently, our model has the ability of identifying and segmenting the salient object and camouflage object well without human annotation.

\subsection{Ablation Study}

\subsubsection{CDP} 

Table \ref{tab:plug} lists the performance of different plug-in models. Thanks to the CDP paradigm, the existing models can be improved further, \textit{e.g.,} DUTS \cite{wang2017learning}, DUT-O \cite{yang2013saliency}, CAMO \cite{le2019anabranch}, CHAMELEON \cite{skurowski2018animal}, COD10K \cite{fan2020camouflaged}, and NC4K \cite{lv2021simultaneously}. Especially our CDP paradigm helps the existing SOD methods get large-margin improvements for the COD task. This indicates that our CDP paradigm has great potentials for the agnostic tasks. Moreover, for the purpose of better describing the contribution of the proposed CDP to the joint learning of SOD and COD, we conduct the experiments of training the existing SOD or COD models using the joint SOD and COD datasets. As shown in Table \ref{tab:cdp_sod} and Table \ref{tab:cdp_cod}, using our CDP paradigm, the joint learning of SOD and COD will be enhanced, compared with the joint learning directly using the joint training datasets.
\subsubsection{Different components} 

Table \ref{tab:ablation}(a) lists the performance of different components. On top of BASE, our IGC gets obvious improvements for both SOD and COD tasks. The performance will be improved further by our CDP, which proves the power of our CDP paradigm for the unified framework of SOD and COD.
\subsubsection{Different initial pseudo masks} 

%
Table \ref{tab:ablation}(b) lists the performance of different initial pseudo masks generations, including TokenCut \cite{wang2022self}, SpectralSeg \cite{melas2022deep}, SelfMask \cite{shin2022unsupervised}, and our pseudo mask. Although SelfMask \cite{shin2022unsupervised} performs well on two SOD benchmarks, our method achieves 27/36 best metrics on SOD/COD tasks. Besides, the update of pseudo labels during training is also crucial, which has been proved in Table \ref{tab:ablationv}(a) with the fact that the pseudo masks update improves the performance a lot for both SOD and COD tasks in the unsupervised setting.
\subsubsection{Different-layer background semantics} 

Table \ref{tab:ablation}(c) lists the performance of different layers, including $\textbf{E}_0 \sim \textbf{E}_4$, of the encoder for background semantics learning within the CDP paradigm. $\textbf{E}_2$ gets all best metrics for SOD/COD, which indicates $\textbf{E}_2$ can well help to extract the background semantics for both salient and camouflaged scenes. Based on this observation, we select $\textbf{E}_2$ to learn the background semantics for SOD and COD.
\subsubsection{Pseudo label}

Following the pipeline of previous unsupervised salient object detection works, we select our initial pseudo labels to train CDP \cite{wu2019cascaded} (SOD method), ZoomNet \cite{pang2022zoom} (COD method) in the unsupervised setting. As shown in Table \ref{tab:ablationv}(b), our approach beats CPD \cite{wu2019cascaded} and ZoomNet \cite{pang2022zoom}. This proves our advantage of the tolerance to coarse pseudo-labels, which comes from the update of pseudo labels during training.

\subsubsection{Training separately.}
We list the results of the proposed method trained using SOD and COD datasets separately in Table \ref{tab:ablationv}(c) and (d) under both supervised and unsupervised settings, respectively. The results of joint training will be lower than that of separate manner, which is reasonable because joint SOD and COD datasets inevitably introduce domain bias.
\begin{figure}[t]
	\centering
	\includegraphics[width=0.96\linewidth]{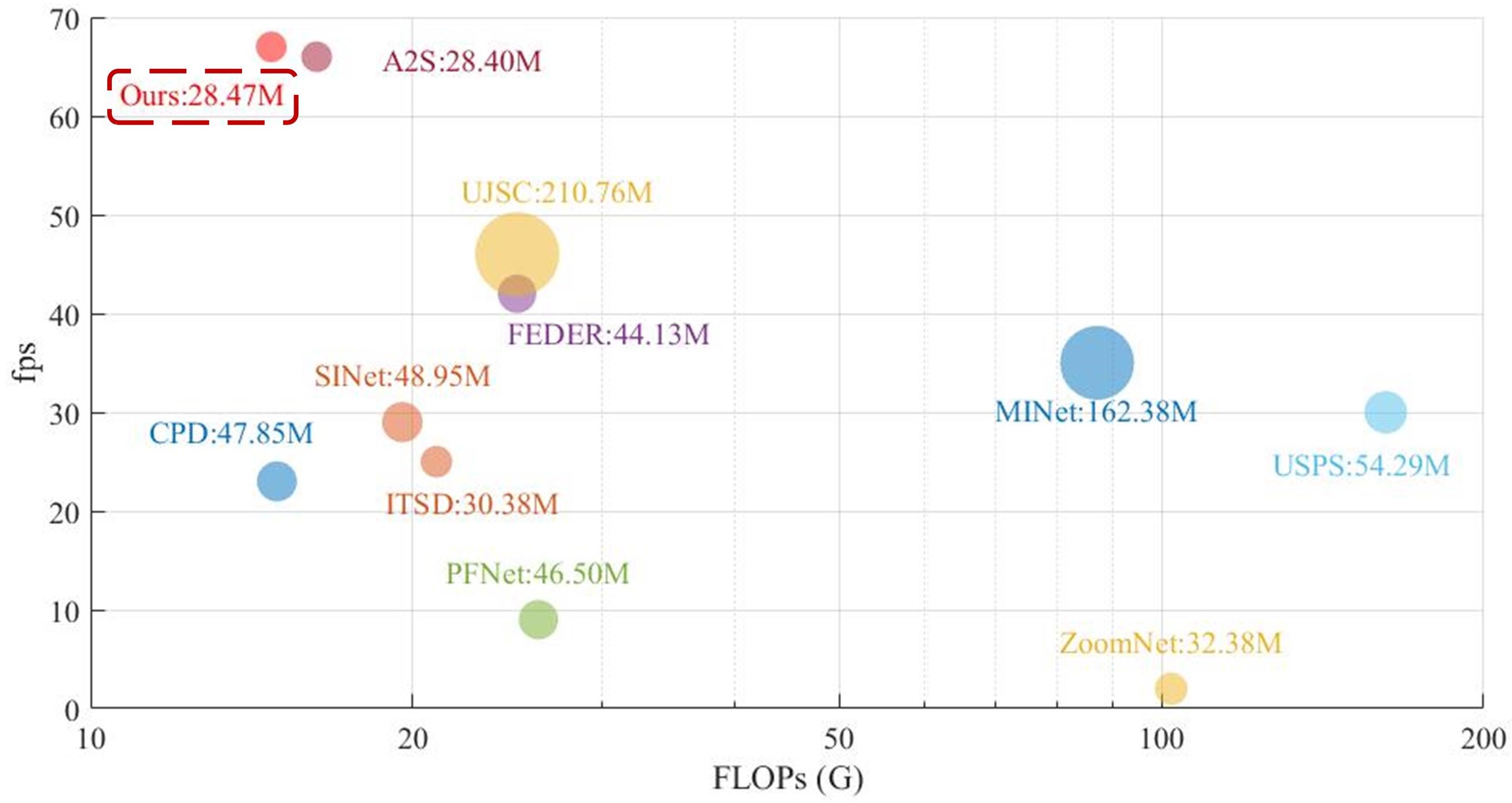}
	\caption{Efficiency comparison.}
	\label{fig: eff}
	\vspace*{-15pt}
\end{figure}

\subsubsection{Inference efficiency}
Fig. \ref{fig: eff} lists parameters, FLOPs, and speed of different methods. It is obvious that our model gets the competitively minimal parameters, least FLOPs, and most fast inference speed of 67 fps, compared with the previous methods, which indicates the potential of our framework for real-time and hardware-friendly applications.

\section{Conclusions}
\label{sec:Conclusion}

In this paper, we have made a task-agnostic unified framework for SOD and COD via a contrastive distillation paradigm based on the agreeable binary segmentation nature. In the supervised setting, our framework performed competitively with the previous task-specific SOD and COD methods. In the unsupervised setting, our framework achieved superior performance on most SOD and COD benchmarks. As well, our work has a real-time inference speed. In the future, we will discover the contribution of our task-agnostic framework for more target-identification tasks, \textit{e.g.}, forgery detection \cite{farid2009image} and shadow removal \cite{sanin2012shadow}, and real-world applications, \textit{e.g.}, lesion detection \cite{yan2018deeplesion} and remote detection \cite{li2020object}.

\section*{Acknowledgment}

This work is supported in part by the National Natural Science Foundation of Jiangsu Province under Grant No. BK20221379, in part by the State Key Laboratory of Reliability and Intelligence of Electrical Equipment under Grant EERI KF2022005, Hebei University of Technology, and in part by the National Natural Science Foundation of China under Grants No. 62293543, No. 62322605 \& No. U21B2048.

\bibliographystyle{cas-model2-names}

\bibliography{cas-refs}



\end{document}